\documentclass[10pt,twocolumn,letterpaper]{article}

\usepackage{iccv}              

\usepackage{multirow}
\usepackage[percent]{overpic}
\usepackage[utf8]{inputenc}

\def\model{Ouroboros}

\definecolor{iccvblue}{rgb}{0.21,0.49,0.74}
\usepackage[pagebackref,breaklinks,colorlinks,allcolors=iccvblue]{hyperref}

\def\paperID{7747} 
\def\confName{ICCV}
\def\confYear{2025}

\title{Ouroboros: Single-step Diffusion Models for \\ Cycle-consistent Forward and Inverse Rendering}

\author{%
  Shanlin Sun$^1\footnotemark[1]$ \quad 
  Yifan Wang$^2\footnotemark[1]$ \quad 
  Hanwen Zhang$^3\footnotemark[1]$ \quad 
  Yifeng Xiong$^1$ \quad 
  Qin Ren$^2$ \\
  Ruogu Fang$^4\footnotemark[2]$ \quad 
  Xiaohui Xie$^1\footnotemark[2]$ \quad 
  Chenyu You$^2\footnotemark[2]$ \\~
\normalsize $^1$ University of California, Irvine \quad
$^2$ Stony Brook University \quad \\
\normalsize $^3$ Huazhong University of Science and Technology \quad
$^4$ University of Florida 
}

\begin{document}
\twocolumn[
    \begin{@twocolumnfalse}
        \maketitle 
        \vspace{-3em}
\begin{center}
    \centering
    \captionsetup{type=figure}
    \includegraphics[width=1.0\linewidth]
    {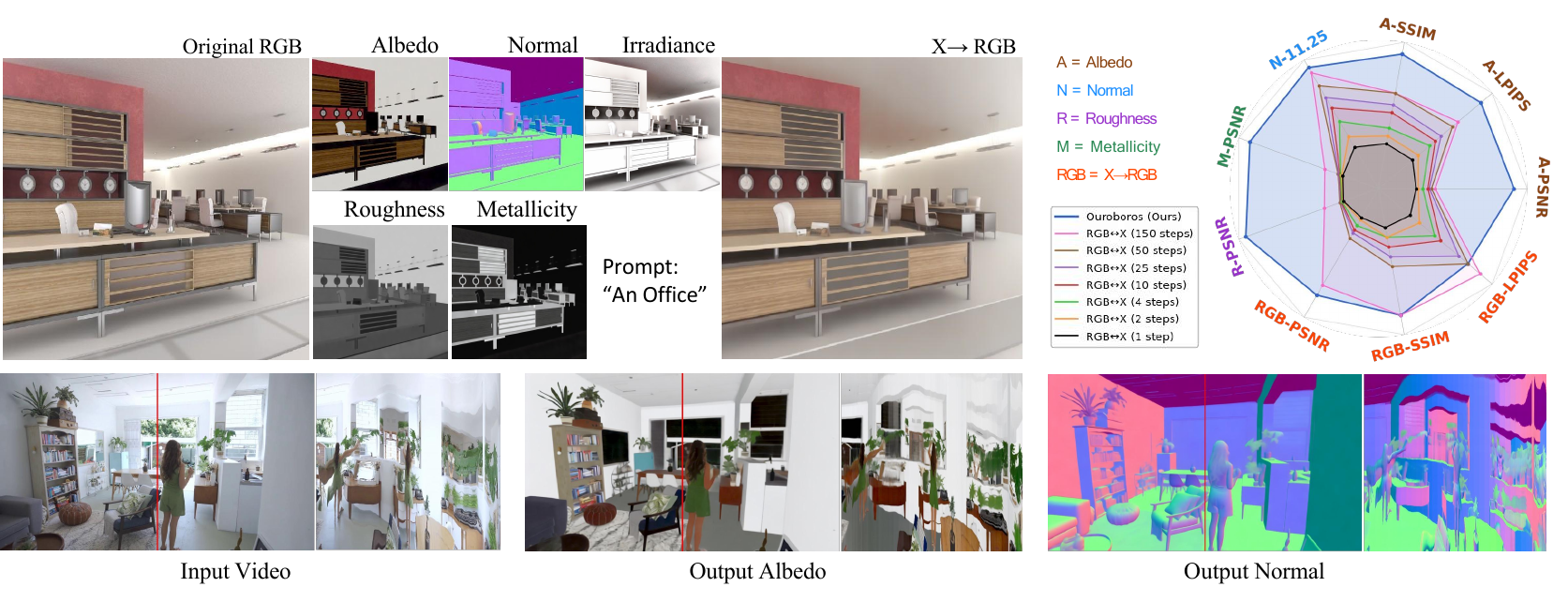}
    \vspace{-2em}
    \captionof{figure}{\textbf{Single-step Diffusion Models for Forward and Inverse Rendering in Cycle Consistency}. \textbf{Left Upper:} \model~ decomposes input images into intrinsic maps (albedo, normal, roughness, metallicity, and irradiance). Given these generated intrinsic maps and textual prompts, our neural forward rendering model synthesizes images closely matching the originals. \textbf{Right Upper:} We extend an end-to-end finetuning technique~\cite{martingarcia2024diffusione2eft} to diffusion-based neural rendering, outperforming state-of-the-art RGB$\leftrightarrow$X~\cite{zeng2024rgb} in both speed and accuracy. The radar plot illustrates numerical comparisons on the InteriorVerse dataset~\cite{zhu2022learning}. \textbf{Bottom:} Our method achieves temporally consistent video inverse rendering without specific finetuning on video data.}
    \label{fig:headline}
\end{center}
    \end{@twocolumnfalse}
]

\renewcommand{\thefootnote}{\fnsymbol{footnote}}
\footnotetext[1]{Equal contribution.}
\footnotetext[2]{Corresponding authors.}

\begin{abstract}
While multi-step diffusion models have advanced both forward and inverse rendering, existing approaches often treat these problems independently, leading to cycle inconsistency and slow inference speed. In this work, we present \textbf{\model}, a framework composed of two single-step diffusion models that handle forward and inverse rendering with mutual reinforcement. Our approach extends intrinsic decomposition to both indoor and outdoor scenes and introduces a cycle consistency mechanism that ensures coherence between forward and inverse rendering outputs. Experimental results demonstrate state-of-the-art performance across diverse scenes while achieving substantially faster inference speed compared to other diffusion-based methods. We also demonstrate that \model can transfer to video decomposition in a training-free manner, reducing temporal inconsistency in video sequences while maintaining high-quality per-frame inverse rendering. Project Page: \href{https://y-research-sbu.github.io/Ouroboros/}{https://y-research-sbu.github.io/Ouroboros/}
\end{abstract}    
\section{Introduction}
\label{sec:intro}

The interdependent processes of forward and inverse rendering are fundamental to computer graphics and vision. Inverse rendering~\cite{barrow1978recovering, ramamoorthi2001signal, barron2014shape} is the problem of estimating geometric, shading, and lighting information from images, a capability essential for applications such as relighting and object insertion. This problem remains challenging due to its inherently under-constrained nature when limited to a single image under single illumination~\cite{grosse2009ground}. On the other hand, forward rendering is to simulate the light transport to render images given the scene's geometry, material, and lighting information. Traditional Physically Based Rendering (PBR)~\cite{pharr2023physically} requires precise geometry and lighting information, which is often difficult to reconstruct accurately through inverse rendering methods.

Recent years have witnessed substantial progress through the development of large-scale annotated synthetic datasets~\cite{li2021openrooms,zhu2022learning,roberts2021hypersim,li2023matrixcity,infinigen2023infinite,infinigen2024indoors} and advanced deep learning techniques. Several data-driven approaches~\cite{li2020inverse,zhu2022irisformer} have emerged to estimate per-pixel intrinsic maps using neural networks, capturing properties such as diffuse color, specular roughness, metallicity, and lighting representations. The advent of diffusion models further benefits inverse rendering capabilities ~\cite{kocsis2023intrinsic,luo2024intrinsicdiffusion,liang2025diffusionrenderer,zeng2024rgb,liang2025diffusionrenderer}, offering powerful priors for estimating ambiguous intrinsic properties and generating photorealistic outputs. More recently, RGB$\leftrightarrow$X~\cite{zeng2024rgb} proposes the first diffusion model for
forward rendering that accommodates flexible combinations of input intrinsic channels. Recently, DiffusionRenderer~\cite{liang2025diffusionrenderer} integrates the idea of RGB$\leftrightarrow$X and Neural Gaffer~\cite{jin2025neural} to video diffusion~\cite{blattmann2023stable}, achieving state-of-the-art performance in video decomposition and relighting. Despite these advances, current diffusion-based approaches exhibit two critical limitations: computational inefficiency and lack of cycle consistency across inverse and forward rendering. 

To this end, we propose \textbf{\model}\footnote{Named after the ancient symbol of a serpent consuming its own tail.}, a unified framework that trains single-step diffusion models for inverse and forward rendering while enforcing cycle consistency between them. Specifically, we demonstrate that a simple end-to-end fine-tuning technique~\cite{martingarcia2024diffusione2eft} can be effectively applied not only to image perception tasks (geometry, material, and lighting estimation) but also to image synthesis operations (forward rendering) while preserving competitive quality.  We fine-tune our single-step inverse and forward rendering models from RGB$\leftrightarrow$X~\cite{zeng2024rgb} using multiple heterogeneous synthetic datasets spanning both interior~\cite{roberts2021hypersim,zhu2022learning} and outdoor datasets~\cite{li2023matrixcity} with varying available intrinsic maps~\cite{zeng2024rgb}. This approach achieves a 50× acceleration in inference speed while maintaining state-of-the-art performance in image decomposition and synthesis.

A significant limitation of independently trained forward and inverse models lies in their inconsistent behavior when applied sequentially, where decomposed properties often fail to accurately reconstruct the original image back. Similar to ControlNet++~\cite{controlnet_plus_plus}, we implement cycle consistency in conditional image understanding and generation. Our single-step generation framework enables straightforward enforcement of cycle consistency in pixel space between inverse and forward rendering during training, similar to CycleGAN~\cite{zhu2017unpaired}. This cycle consistency mechanism facilitates the incorporation of unannotated real-world data into the training process through self-supervision, thereby reducing dependence on large-scale, high-quality, and diverse synthetic renderings with paired annotations.

Beyond its primary capabilities, \model offers valuable benefits for downstream applications. We explore a simple, training-free approach to extend our image-based neural inverse rendering method to consistent long video intrinsic decomposition by flattening spatial-temporal video patches and extending 2D convolution kernels into pseudo-3D kernels. Additionally, we demonstrate promising preliminary results in fine-tuning \model for single-step diffusion-based object removal and insertion.

In summary, our contributions include:
\begin{itemize}
    \item A state-of-the-art fast diffusion-based neural framework for inverse and forward rendering, validated across indoor and outdoor scene domains;
    \item A cycle consistency training methodology that ensures coherence between image decomposition and synthesis while enabling the utilization of heterogeneous synthetic datasets and unannotated real-world data;
    \item A training-free approach for achieving temporal stability in video applications despite training exclusively on image data.
\end{itemize}

\section{Related Works}
\label{sec:related}
\paragraph{Diffusion Models for Image Understanding.}
Diffusion models~\cite{ho2020denoising,ho2022classifier} excel at generating photorealistic images by reversing a learned noising process. Conditioned on inputs like text prompts~\cite{rombach2022high,saharia2022photorealistic,esser2024scaling,blattmann2023stable}, they have advanced numerous vision tasks, including conditional generation~\cite{zhang2023adding,ye2023ip,mou2024t2i,qin2023unicontrol}, image and video editing~\cite{brooks2023instructpix2pix,kawar2023imagic,yang2023paint,chen2024anydoor,ceylan2023pix2video,liu2024video,chai2023stablevideo}, and story generation~\cite{wang2024autostory,zhou2025storydiffusion}.
Beyond generation, diffusion models are further adapted for perception tasks like geometry estimation~\cite{ke2024repurposing,fu2024geowizard,gui2024depthfm,ye2024stablenormal,he2024lotus}, semantic segmentation~\cite{xu2024matters,zhu2024unleashing}, and pose estimation~\cite{gong2023diffpose,wang2024lavin,le2024one}. By reformulating these as conditional generation problems, pre-trained models can be fine-tuned to produce dense prediction maps from single images, leveraging their capacity for intricate detail.
Fine-tuning pre-trained diffusion models has also become an effective strategy for achieving various image editing effects, including altering lighting conditions. DiLightNet~\cite{zeng2024dilightnet} provides fine-grained control over lighting during image generation by using radiance hints to guide the diffusion process. LightIt~\cite{kocsis2024lightit} proposes an identity-preserving relighting model conditioned on an image and a target shading. Relightful Harmonization~\cite{ren2024relightful} manipulates the illumination of foreground objects using background conditions. Similarly, Neural Gaffer~\cite{jin2025neural} relights foregrounds with target environment maps. More recently, IC-Light~\cite{zhangscaling} focuses on scaling up the training of diffusion-based illumination editing models by imposing consistent light transport, ensuring illumination is modified while preserving other intrinsic image properties. These approaches often leverage techniques such as 3D rendering, Neural Radiance Fields (NeRF), and synthetic data to achieve sophisticated illumination and appearance control.
\paragraph{Intrinsic Decomposition} as defined by~\cite{barrow1978recovering}, separates an RGB image into components like albedo and irradiance, with modern methods also estimating factors such as roughness and normals. To improve accuracy, these methods leverage diverse inputs including human annotations~\cite{kovacs2017shading,wu2023measured,zhou2015learning}, ordinal cues~\cite{careaga2023intrinsic,zoran2015learning}, physical priors~\cite{ramanagopal2024theory,yoshida2023light}, structural models~\cite{geng2023tree}, and multi-view data~\cite{philip2019multi,ye2023intrinsicnerf,wu2024deferredgs}.
Current approaches also utilize pretrained generative models~\cite{karras2019style,rombach2022high} for extracting intrinsic images, either via latent space optimization~\cite{bhattad2023stylegan}, low-rank adaptations~\cite{du2023generative}, or image-conditioned diffusion generative models~\cite{kocsis2023intrinsic,luo2024intrinsicdiffusion}. The advancement of these models is coupled with the growing availability of high-quality, large-scale synthetic datasets, including interior datasets like InteriorNet~\cite{InteriorNet18}, OpenRooms~\cite{li2021openrooms}, InteriorVerse~\cite{zhu2022learning}, and Hypersim~\cite{roberts2021hypersim}, as well as outdoor datasets such as MatrixCity~\cite{li2023matrixcity}.
The state-of-the-art method, RGB$\leftrightarrow$X~\cite{zeng2024rgb}, estimates multiple intrinsic buffers using synthetic interior data. Our work builds on this foundation by fine-tuning with both interior and outdoor datasets to achieve faster, more accurate decomposition. Recently, DiffusionRenderer~\cite{liang2025diffusionrenderer} extends similar diffusion-based inverse rendering techniques to the video domain.
\paragraph{Neural Image Synthesis from Decompositions.}
While state-of-the-art rendering uses Monte Carlo simulation~\cite{pharr2023physically}, synthesizing images from intrinsic decompositions is challenging. Traditional methods require full 3D geometry and explicit light/material properties, which are absent in decomposed representations.
To address this, recent work leverages neural networks to synthesize images from intrinsic buffers. Approaches include using CNNs for screen-space shading~\cite{nalbach2017deep}, employing screen-space ray methods~\cite{zhu2022learning}, and using diffusion models to compose realistic images from intrinsic channels~\cite{zeng2024rgb}. Our work aligns with this direction, using intrinsic maps to guide image synthesis while enforcing cycle consistency between decomposition and synthesis.
Neural image relighting methods have been developed with explicit decomposition~\cite{griffiths2022outcast,pandey2021total,yu2020self,kocsis2024lightit,kim2024switchlight} or implicit representations~\cite{rudnev2022nerf,wang2023neural,lin2023urbanir}. DiffusionRenderer~\cite{liang2025diffusionrenderer} takes pre-defined environment maps, instead of diffuse shading maps, as inputs to the forward rendering model, thereby supporting zero-shot image relighting by incorporating novel environment maps. ZeroComp~\cite{zhang2024zerocomp} and RGB$\leftrightarrow$X~\cite{zeng2024rgb} both demonstrate 3D object compositing methods built upon neural rendering. 
\section{Preliminary: RGB$\leftrightarrow$X}
\label{sec:pre}
RGB$\leftrightarrow$X~\cite{zeng2024rgb} proposed a unified diffusion framework for both inverse rendering (RGB→X) and forward rendering (X→RGB) using latent diffusion models~\cite{rombach2022high} that operate in VAE~\cite{kingma2013auto} latent space with encoder $\mathcal{E}$ and decoder $\mathcal{D}$. Their approach operates on five intrinsic channels: normal vector $\mathbf{n} \in \mathbb{R}^{H \times W \times 3}$, albedo $\mathbf{a} \in \mathbb{R}^{H \times W \times 3}$, roughness $\mathbf{r} \in \mathbb{R}^{H \times W}$, metallicity $\mathbf{m} \in \mathbb{R}^{H \times W}$, and diffuse irradiance $\mathbf{E} \in \mathbb{R}^{H \times W \times 3}$. The RGB→X model estimates intrinsic channels from an RGB image $\mathbf{I} \in \mathbb{R}^{H \times W \times 3}$ by fine-tuning a pre-trained latent diffusion model with v-prediction~\cite{salimans2022progressive}:
\begin{equation}
\mathbf{v}_t^{RGB \rightarrow X} = \sqrt{\bar{\alpha}_t}\epsilon - \sqrt{1 - \bar{\alpha}_t}\mathbf{z}_0^X,
\end{equation}
where $t$ is the diffusion time step, $\epsilon \sim \mathcal{N}(\mathbf{0}, \mathbf{I})$ is Gaussian noise, $\bar{\alpha}_t$ is a scalar function of $t$, and $\mathbf{z}_0^X$ is the clean target latent encoding the intrinsic channels. To handle multiple output channels with a single model, they repurpose the text prompt as a switch mechanism, using fixed prompts (e.g., normal, albedo, roughness) to control which intrinsic channel is produced.
The X→RGB model synthesizes RGB images from intrinsic channels, defining the clean target latent as $\mathbf{z}_0^{RGB} = \mathcal{E}(\mathbf{I})$. The model also employs v-prediction:
\begin{equation}
\mathbf{v}_t^{X \rightarrow RGB} = \sqrt{\bar{\alpha}_t}\epsilon - \sqrt{1 - \bar{\alpha}_t}\mathbf{z}_0^{RGB},
\end{equation}
and uses a channel dropout strategy during training to handle heterogeneous datasets:
\begin{equation}
\mathbf{z}_t^X = (\mathcal{P}(\mathbf{n}), \mathcal{P}(\mathbf{a}), \mathcal{P}(\mathbf{r}), \mathcal{P}(\mathbf{m}), \mathcal{P}(\mathbf{E})),
\end{equation}
where $\mathbf{z}_t^X$ represents the noisy latent at time step $t$, and $\mathcal{P}(x) \in {\mathcal{E}(x), 0}$ is the dropout function. This approach allows generation from any subset of channels at inference time while maintaining a single unified model. Notably, irradiance $\mathbf{E}$ is handled differently than other intrinsic channels in the X→RGB model; while normal, albedo, roughness, and metallicity maps are encoded through the full-resolution encoder $\mathcal{E}$, the irradiance is instead directly downsampled to latent resolution.
\section{Method}
\begin{figure*}
  \centering
   \includegraphics[width=1.0\linewidth]{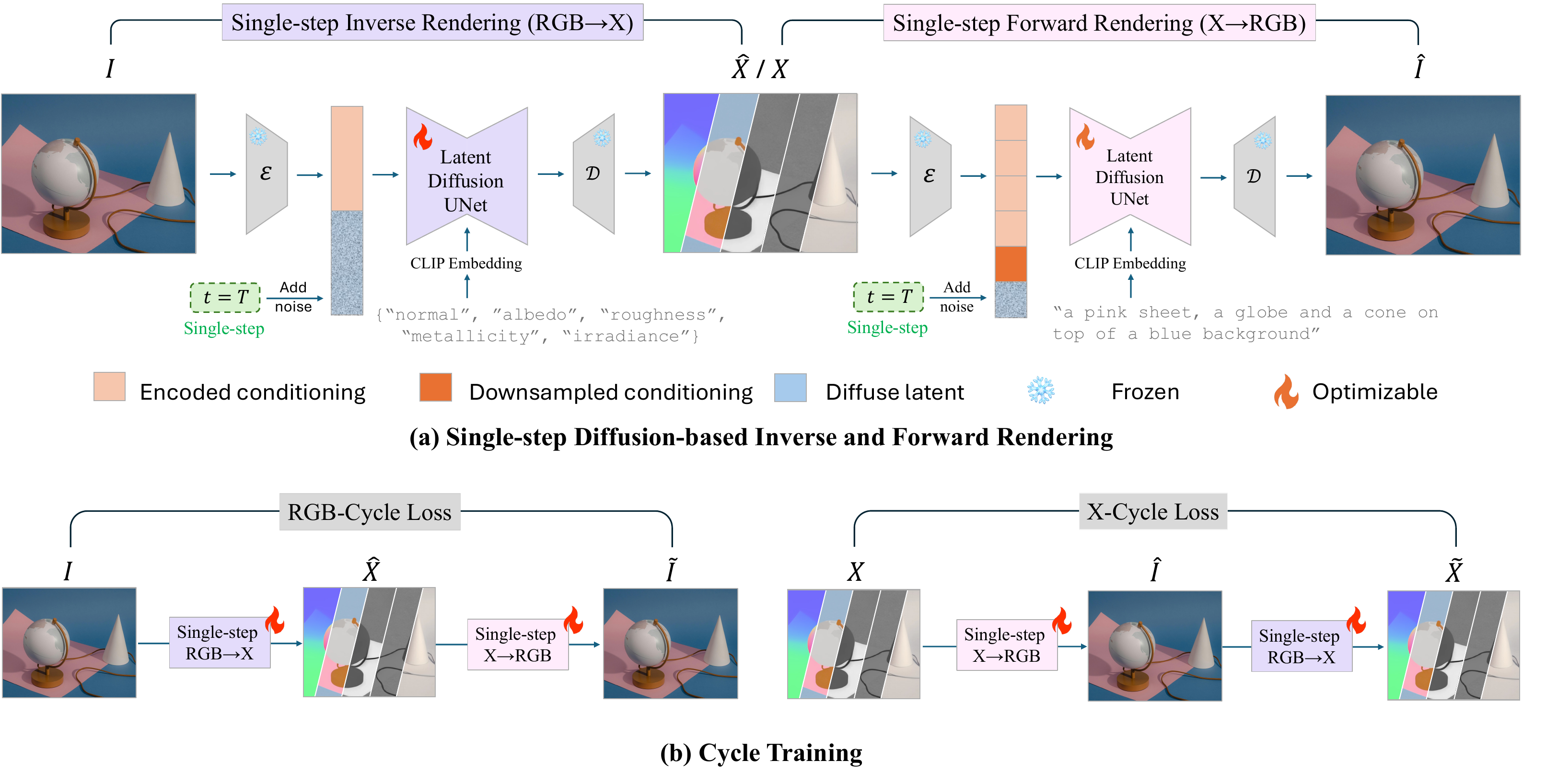} 
\vspace{-2em}
\caption{\textbf{Overview of \model~ Pipeline.} (a) presents the training pipeline of our single-step Diffusion-based inverse and forward rendering model. For inverse rendering, the model takes the image $I$ and text prompt indicating the output intrinsic maps as input to finetune the latent diffusion UNet. For forward rendering, the model is fed with concatenated intrinsic maps along with simple image description to estimate the original image. (b) provides the overview of cycle training pipeline.}
   \label{fig:framework}
\end{figure*}

\model is composed of two single-step diffusion models serving for inverse and forward rendering respectively. As illustrated in Fig.~\ref{fig:framework}, the neural inverse rendering (RGB→X) predicts pixel-wise geometry, material and lighting properties from an input image. The neural forward rendering is to syntheize image from intrinsic buffers.

We describe our single-step diffusion-based inverse and forward rendering finetuning strategy in Sec.~\ref{subsec:finetune} as well as cycle training pipeline in Sec.~\ref{subsec:cycle}, and discuss how we inference video data with our image-based model in Sec.~\ref{subsec:video}.

\label{sec:method}
\subsection{Finetuning Single-Step Prediction Model}
\label{subsec:finetune}

Inspired by E2E~\cite{martingarcia2024diffusione2eft}, we finetune pre-trained RGB$\leftrightarrow$X diffusion models to generate high-quality intrinsic maps with a single-step inference. Given the pairwise data ($X$, $I$), where $I$ denotes the RGB image and $X$ represents the set of intrinsic maps, the diffusion model uses one as a conditional input to generate the other. In our finetuning framework, most diffusion modules are frozen except for the UNet.

\paragraph{Finetuning pipeline.} To enable efficient single-step prediction, we fix the timestep to $t = T$ during training, forcing the model to learn to denoise from the most noisy state to the target in a single step. Unlike E2E~\cite{martingarcia2024diffusione2eft}, which uses zero noise as the initial state, we apply multi-resolution noise at timestep $T$ to the target latent, which means our single-step model is not deterministic. This non-deterministic approach is particularly appropriate for intrinsic decomposition, which inherently admits multiple possible solutions unlike tasks such as depth or normal estimation that have more definitive ground truths. 

During training, the UNet output is converted into a latent prediction using the \(\mathbf{v}\)-parameterization~\cite{salimans2022progressive}:
\begin{equation}
\hat{\mathbf{z}}_0 = \sqrt{\overline{\alpha}_T}\mathbf{z}_T - \sqrt{1 - \overline{\alpha}_T}\hat{\mathbf{v}}_{\theta},
\end{equation}
where $\hat{\mathbf{z}}_0$ is the predicted denoised latent, $\mathbf{z}_T$ is the noised input, and $\hat{\mathbf{v}}_{\theta}$ is the diffusion Unet's output. This predicted latent is subsequently decoded into the original space via the VAE decoder for comparison with the ground truth.

\paragraph{Task-specific loss functions.} We employ different loss functions tailored to each type of intrinsic map:

For normal predictions, we use a loss based on the angular difference between estimated and ground-truth normals:
\begin{equation}
\mathcal{L}_{\mathbf{n}} = \frac{1}{N}\sum_{i}\text{arccos}\frac{\mathbf{n}_i \cdot \hat{\mathbf{n}}_i }{||\mathbf{n}_i|| \cdot ||\hat{\mathbf{n}}_i||},
\end{equation}
where $N$ represents the total number of pixels, $\mathbf{n}_i$ denotes ground-truth normal vectors, and $\hat{\mathbf{n}}_i$ represents predicted normal vectors.

For irradiance prediction, we apply an affine-invariant loss function~\cite{ranftl2020towards}:
\begin{equation}
\mathcal{L}_{\mathbf{E}} = |\mathbf{E} - \mathbf{S}\hat{\mathbf{E}} - \mathbf{T}|_F^2,
\end{equation}
where $\mathbf{E}, \hat{\mathbf{E}}$ are ground truth and predicted irradiance maps, $\mathbf{S}$ is a diagonal scale matrix, and $\mathbf{T}$ contains channel-specific shift values. Parameters are determined through least-square fitting for each channel independently, accommodating the ambiguity in decomposing images into albedo and irradiance.

For all other maps, including RGB, albedo, roughness, and metallicity, we utilize the Mean Squared Error (MSE):
\begin{equation}
\mathcal{L}_{\{a,r,m,RGB\}} = \frac{1}{N}\sum_i|\textbf{y}_i - \hat{\textbf{y}}_i|_F^2,
\end{equation}
where $y_i$ represents ground-truth pixel values and $\hat{y}_i$ corresponds to predicted values. For inverse rendering task, the final loss is computed as the sum of these individual intrinsic map specific loss functions. For forward rendering task, the loss is $\mathcal{L}_{RGB}$.

\paragraph{Training Data.} Due to the limited availability of datasets containing complete RGB-X pairs, we integrate multiple complementary datasets in our training pipeline. For indoor scenes, we utilize Hypersim~\cite{roberts2021hypersim} and InteriorVerse~\cite{zhu2022learning}, following the heterogeneous data handling approach in~\cite{zeng2024rgb} that employs channel dropout to accommodate varying availability of intrinsic channels across datasets. To ensure robustness to outdoor scenes, we incorporate MatrixCity~\cite{li2023matrixcity}, a large-scale urban dataset with photorealistic images, each accompanied by normal, albedo, metallic, and roughness maps. We randomly sample 17,000 image (with paired intrinsic maps) from each dataset.

For the forward rendering model (X→RGB), which requires image descriptions as conditions, we employ BLIP-2~\cite{li2023blip} for indoor scenes and BLIP~\cite{li2022blip} for outdoor scenes. Empirically, we find that BLIP-2 provides more detailed annotations suitable for complex indoor environments, while BLIP generates more concise and accurate descriptions for outdoor scenes.
\subsection{Cycle Training}
\label{subsec:cycle}
After an initial round of finetuning, we obtain two complementary diffusion models capable of single-step inference: one for inverse rendering (RGB→X) and another for forward rendering (X→RGB). However, as these models are trained independently, they exhibit deficiencies in cycle consistency when applied sequentially. To address this limitation, we implement a cycle-consistent training approach similar to CycleGAN~\cite{zhu2017unpaired}.

Given an input pair ($\mathbf{X}$, $\mathbf{I}$), where $\mathbf{X}$ represents intrinsic maps and $\mathbf{I}$ denotes the RGB image, we first utilize our pre-trained single-step models to generate the corresponding outputs ($\hat{\mathbf{I}}$, $\hat{\mathbf{X}}$). Subsequently, we use these generated outputs as inputs for a second inference pass, producing ($\tilde{\mathbf{X}}$, $\tilde{\mathbf{I}}$). This enables us to define cycle consistency losses:
\begin{equation}
\mathcal{L}_{cycle} = \mathcal{L}_{\textbf{X} \rightarrow \hat{\textbf{I}} \rightarrow \tilde{\textbf{X}}} + \mathcal{L}_{\textbf{I} \rightarrow \hat{\textbf{X}} \rightarrow \tilde{\textbf{I}}} = |\mathbf{X} - \tilde{\mathbf{X}}|^2 + |\mathbf{I} - \tilde{\mathbf{I}}|^2
\end{equation}
During this additional finetuning phase, we optimize both models jointly using a combination of the task-specific losses from Section~\ref{subsec:finetune} and the cycle consistency loss. This approach serves two important purposes: (1) it improves bidirectional consistency between the inverse and forward rendering processes, and (2) it helps mitigate data scarcity issues in training the forward rendering model by leveraging the cycle structure.

\paragraph{Training Data.} In addition to the annotated datasets mentioned in Section~\ref{subsec:finetune}, we leverage 20,000 images sampled from MSCOCO~\cite{lin2014microsoft} and Flickr30k~\cite{plummer2015flickr30k} datasets for cycle training. These natural image collections provide diverse visual content that helps enhance the generalization capabilities of our models.
\subsection{Video Inference}
\label{subsec:video}
\begin{figure}[ht]
\centering
\includegraphics[width=1.0\linewidth]{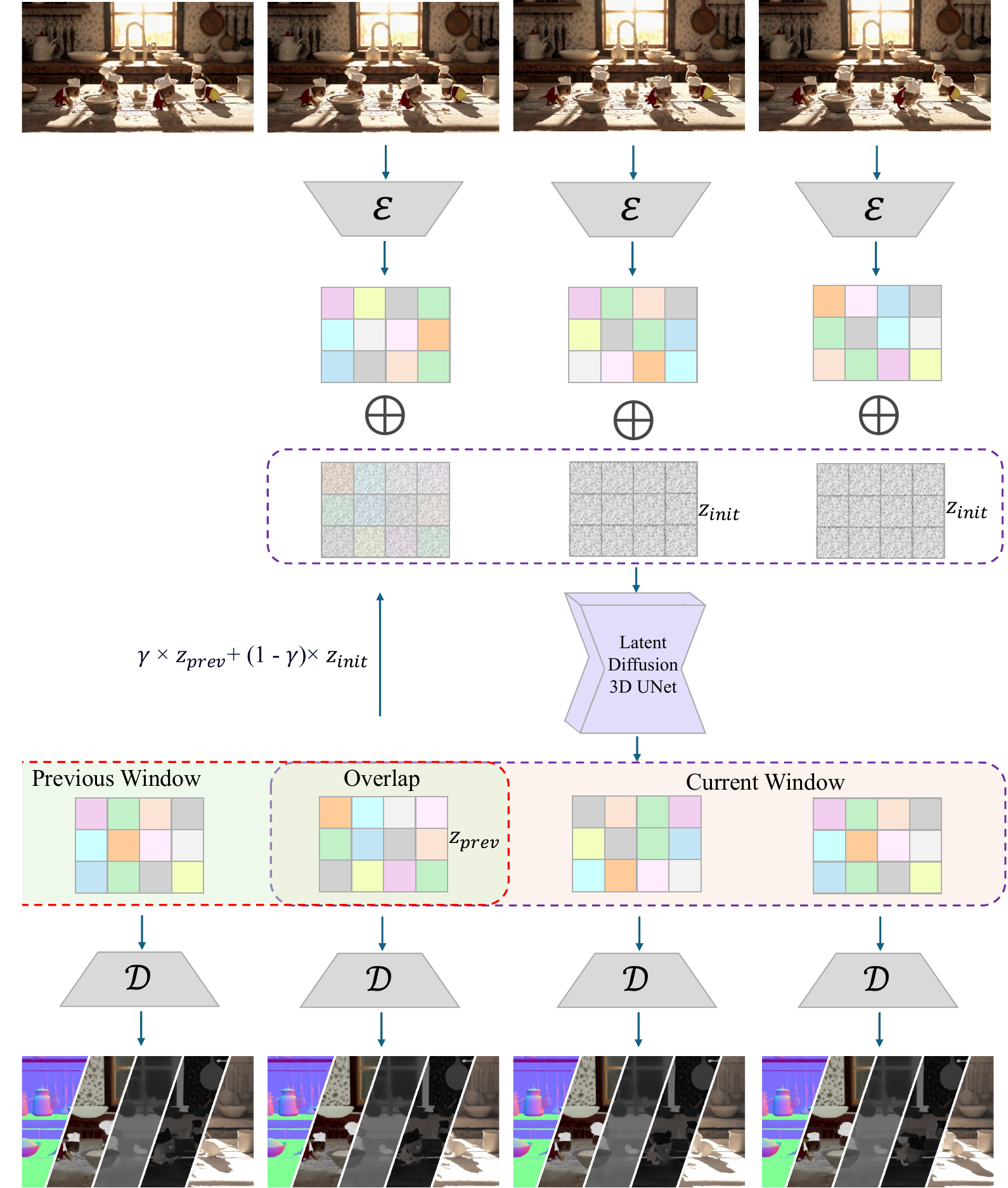}
\caption{\textbf{Iterative Video Generation Pipeline.} Overlapping windows are processed sequentially, with latent representations from previous windows guiding the initialization of overlapping regions. In practice, the window size and overlap are larger than the figure shown.}
\label{fig:video_inference_framework}
\end{figure}
For video inference, although training a native video diffusion model is natural, it typically requires significantly larger datasets, higher computational costs, and longer training times. Instead, we leverage our pretrained 2D diffusion model without any fine-tuning to achieve video generation capabilities.

Naive per-frame application of 2D diffusion models often results in temporal discontinuities and flickering artifacts due to the absence of inter-frame dependencies. To address this limitation, we extend our 2D architecture to handle temporal information effectively. Inspired by prior works such as VDM \cite{ho2022video} and FLATTEN \cite{cong2024flatten}, we extend 2D convolution layers into a pseudo-3D architecture by replacing $3 \times 3$ kernels with $1 \times 3 \times 3$ kernels. Furthermore, our approach flattens patches from multiple frames and applies attention mechanisms across both spatial and temporal dimensions, improving the coherence of generated videos.

However, directly processing an entire video as input and generating the full video output in a single pass poses significant challenges due to GPU memory constraints. To overcome this limitation, we adopt an iterative inference strategy that processes videos in manageable segments. Specifically, we divide the video into overlapping windows with a fixed stride, where each segment is processed independently using our pseudo-3D diffusion model. 

To maintain temporal consistency across segments, we employ a technique inspired by Lotus \cite{he2024lotus}. We take the predicted latent $\mathbf{z}_{\text{prev}}$ of the overlap region from the previous window and apply a weighted combination with noise $\boldsymbol{\epsilon}$ using a pre-defined scale $\gamma$. The result serves as the initial latent input for the overlapping region in the next iteration:
\begin{equation}
\mathbf{z}_{\text{init}} = \gamma \cdot \mathbf{z}_{\text{prev}} + (1- \gamma) \cdot \boldsymbol{\epsilon}
\end{equation}
where we empirically set $\gamma = 0.1$ in our experiments. This approach ensures smooth transitions between video segments while maintaining computational efficiency. See supplementary material for a detailed demonstration of our video inference pipeline.

\section{Experiment}
\label{sec:exp}

\begin{table*}[!htbp]
    \centering
    \caption{\textbf{Albedo Prediction Results.} $\uparrow$ ($\downarrow$) means that the higher (lower), the better. We highlight the best results in \textbf{bold} and the second best with \underline{underlined} format.}
    \vspace{-0.5em}
    \resizebox{\textwidth}{!}{
    \begin{tabular}{@{}l|cccc|cccc|cccc}
        \toprule
        \multirow{2}*{\textbf{Method}}  & \multicolumn{4}{c|}{\textbf{Hypersim}~\cite{roberts2021hypersim}} & \multicolumn{4}{c|}{\textbf{MatrixCity}~\cite{zhu2022learning}} & \multicolumn{4}{c}{\textbf{InteriorVerse}~\cite{li2023matrixcity}} \\
        \cmidrule(lr){2-5} \cmidrule(lr){6-9} \cmidrule(lr){10-13}
        
        & \textbf{PSNR$\uparrow$} & \textbf{LPIPS$\downarrow$} & \textbf{SSIM$\uparrow$} & \textbf{RMSE$\downarrow$} &  \textbf{PSNR$\uparrow$} & \textbf{LPIPS$\downarrow$} & \textbf{SSIM$\uparrow$} & \textbf{RMSE$\downarrow$} & \textbf{PSNR$\uparrow$} & \textbf{LPIPS$\downarrow$} & \textbf{SSIM$\uparrow$} & \textbf{RMSE$\downarrow$}  \\
        \midrule
        RGB$\leftrightarrow$X~\cite{zeng2024rgb} & \underline{18.67} & \textbf{0.20} & \underline{0.59} & \underline{0.43} &  12.61 & 0.26 & 0.53 & 0.50 & 16.17 & \underline{0.17} & 0.77 & 0.30 \\
        \citet{zhu2022learning} & 11.76 & 0.45 & 0.51 & 0.47 & 16.11 & 0.47 & 0.59 & 0.33 & \underline{17.19} & 0.22 & \underline{0.81} & \underline{0.29} \\
        \citet{kocsis2023intrinsic} & 12.40 & 0.31 & 0.57 & 0.49  & 15.66 & 0.36 & 0.57 & 0.34& 14.62 & 0.21 & 0.78 & 0.37 \\
        IntrinsicAnything\cite{chen2024intrinsicanything} & 10.39 & 0.51 & 0.50 & 0.55  & 15.62 & 0.49 & 0.57 & 0.55 & 13.12 & 0.32 & 0.72 & 0.43 \\
        \citet{careaga2024colorful} & 12.01 & 0.31 & 0.57 & 0.45  & \underline{17.30} & \underline{0.21} & \underline{0.71} & \underline{0.28} &  15.51 & 0.21 & 0.80 & 0.32 \\
        \textbf{Ours} & \textbf{18.98} & \underline{0.23} & \textbf{0.71} & \textbf{0.17}  & \textbf{25.38} & \textbf{0.17} & \textbf{0.77} & \textbf{0.11} & \textbf{22.07} & \textbf{0.12} & \textbf{0.87} & \textbf{0.17} \\
        \bottomrule
    \end{tabular}
    }
    \label{albedo prediction}
\end{table*}
\begin{table}[!htbp]
    \centering
    \caption{Normal Prediction Results.}
    \vspace{-0.5em}
    \resizebox{\columnwidth}{!}{
    \begin{tabular}{@{}l|cc|cc|cc}
        \toprule
        \multirow{2}*{\textbf{Method}}  & \multicolumn{2}{c|}{\textbf{Hypersim}} & \multicolumn{2}{c|}{\textbf{MatrixCity}} & \multicolumn{2}{c}{\textbf{InteriorVerse}} \\
        \cmidrule(lr){2-3} \cmidrule(lr){4-5} \cmidrule(lr){6-7}
        
        & \textbf{Mean$\downarrow$} & \textbf{11.25°$\uparrow$} & \textbf{Mean$\downarrow$} & \textbf{11.25°$\uparrow$} & \textbf{Mean$\downarrow$} & \textbf{11.25°$\uparrow$}  \\
        \midrule
        RGB$\leftrightarrow$X~\cite{zeng2024rgb} & 17.21 & 73.93 & 23.82 & 49.74 & 12.10 & 80.03 \\
        \citet{zhu2022learning} & 53.76 & 10.57 & 35.00 & 17.57 & 17.16 & 64.90 \\
        Lotus~\cite{he2024lotus} & 17.61 & 71.38 & 25.06 & 62.34 & 15.38 & 62.12 \\
        StableNormal\cite{ye2024stablenormal} & 16.65 & \underline{75.51} & 18.18 & 63.28 & \underline{10.73} & \underline{82.13} \\
        E2E~\cite{martingarcia2024diffusione2eft} & \underline{16.30} & 74.00 & \textbf{13.91} & \underline{67.09} & 15.87 & 57.30 \\
        \textbf{Ours} & \textbf{11.98} & \textbf{80.71} & \underline{18.12} & \textbf{76.21} &  \textbf{9.58} & \textbf{83.54} \\
        \bottomrule
    \end{tabular}
    }
    \label{normal prediction}
\end{table}

\begin{table*}[!htbp]
    \centering
    \begin{minipage}[t]{0.3\textwidth}
        \centering
        \caption{Irradiance Prediction Results.}
        \vspace{-0.5em}
        \resizebox{\columnwidth}{!}{
        \begin{tabular}{@{}l|cc}
            \toprule
            \multirow{2}*{\textbf{Method}}  & \multicolumn{2}{c}{\textbf{Hypersim}} \\
            \cmidrule(lr){2-3}
            & \textbf{PSNR$\uparrow$} & \textbf{LPIPS$\downarrow$}  \\
            \midrule
            RGB$\leftrightarrow$X & 11.64 & \textbf{0.23} \\
            \citet{du2023generative} & 9.51 & 0.56 \\
            \textbf{Ours} & \textbf{12.07} & \underline{0.29} \\
            \bottomrule
        \end{tabular}
        }
        \label{tab:irradiance_prediction}
    \end{minipage}
    \hfill
    \begin{minipage}[t]{0.665\textwidth}
        \centering
        \caption{Roughness and Metallicity Prediction Results.}
        \vspace{-0.5em}
        \resizebox{\columnwidth}{!}{
        \begin{tabular}{@{}l|cccc|cccc}
            \toprule
            & \multicolumn{4}{c}{\textbf{MatrixCity}} & \multicolumn{4}{c}{\textbf{InteriorVerse}}  \\
            \cmidrule(lr){2-5} \cmidrule(lr){6-9}
            \textbf{Method}  & \multicolumn{2}{c}{\textbf{Roughness}} & \multicolumn{2}{c}{\textbf{Metallicity}}  & \multicolumn{2}{c}{\textbf{Roughness}} & \multicolumn{2}{c}{\textbf{Metallicity}} \\
            \cmidrule(lr){2-3} \cmidrule(lr){4-5} \cmidrule(lr){6-7} \cmidrule(lr){8-9} 
            & \textbf{PSNR$\uparrow$} & \textbf{LPIPS$\downarrow$} & \textbf{PSNR$\uparrow$} & \textbf{LPIPS$\downarrow$} & \textbf{PSNR$\uparrow$} & \textbf{LPIPS$\downarrow$} & \textbf{PSNR$\uparrow$} & \textbf{LPIPS$\downarrow$} \\
            \midrule
            RGB$\leftrightarrow$X~\cite{zeng2024rgb} & \underline{23.82} & \underline{0.3655} & \underline{6.83} & 0.57 & \underline{12.07} & 0.35 & 8.04 & \underline{0.45} \\
            \citet{zhu2022learning} & 7.0028 & 0.6024 & 4.87 & \underline{0.68} & 7.51 & 0.51 & 6.45 & 0.78 \\
            \citet{kocsis2023intrinsic} & 8.4766 & 0.4419 & 10.67 & 0.38 & 11.29 & \underline{0.32} & \underline{8.93} & 0.71 \\
            \textbf{Ours} & \textbf{24.04} & \textbf{0.2301} & \textbf{26.32} & \textbf{0.14} & \textbf{17.83} & \textbf{0.08} & \textbf{13.85} & \textbf{0.12} \\
            \bottomrule
        \end{tabular}
        }
        \label{tab:roughness_metal}
    \end{minipage}
\end{table*}
\begin{figure*}[!htbp]
    \centering
\begin{minipage}{\linewidth}
    \begin{minipage}{\linewidth}
        \centering
        \subfloat[
        \textbf{Albedo Estimation.} 
        Our method outperforms that of \citet{kocsis2023intrinsic} and \citet{careaga2024colorful} in estimation quality. Although RGB$\leftrightarrow$X~\cite{zeng2024rgb} provides estimation with clear detail, its color is adversely impacted by the presence of light. Only our model achieves both image quality and correct color space.
        ]{
            \begin{overpic}[width=0.16\linewidth]{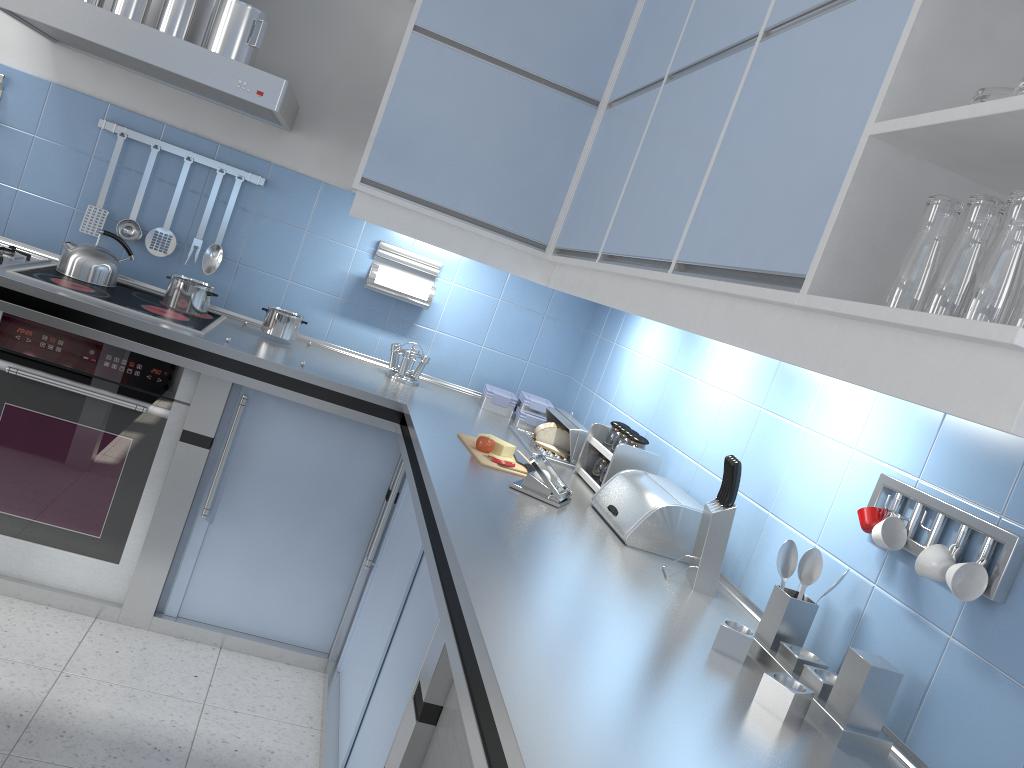}
                \put(25,79){\small Input image}
            \end{overpic}
            \hfill
            \begin{overpic}[width=0.16\linewidth]{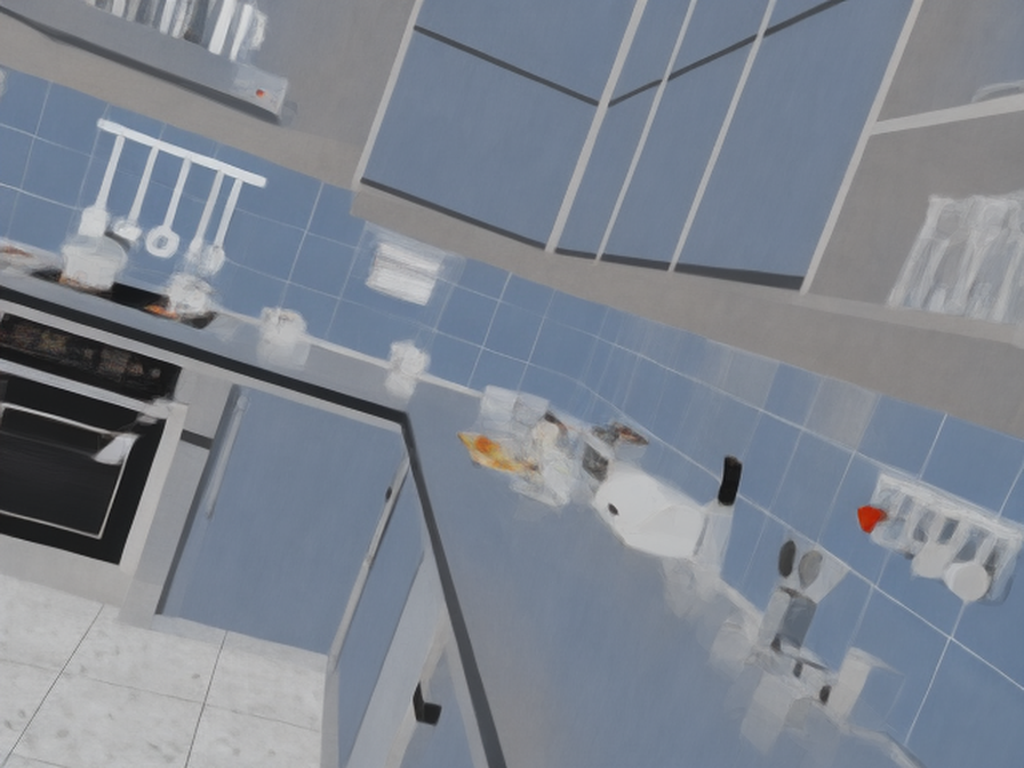}
                \put(11,79){\small\citet{kocsis2023intrinsic}}
            \end{overpic}
            \hfill
            \begin{overpic}[width=0.16\linewidth]{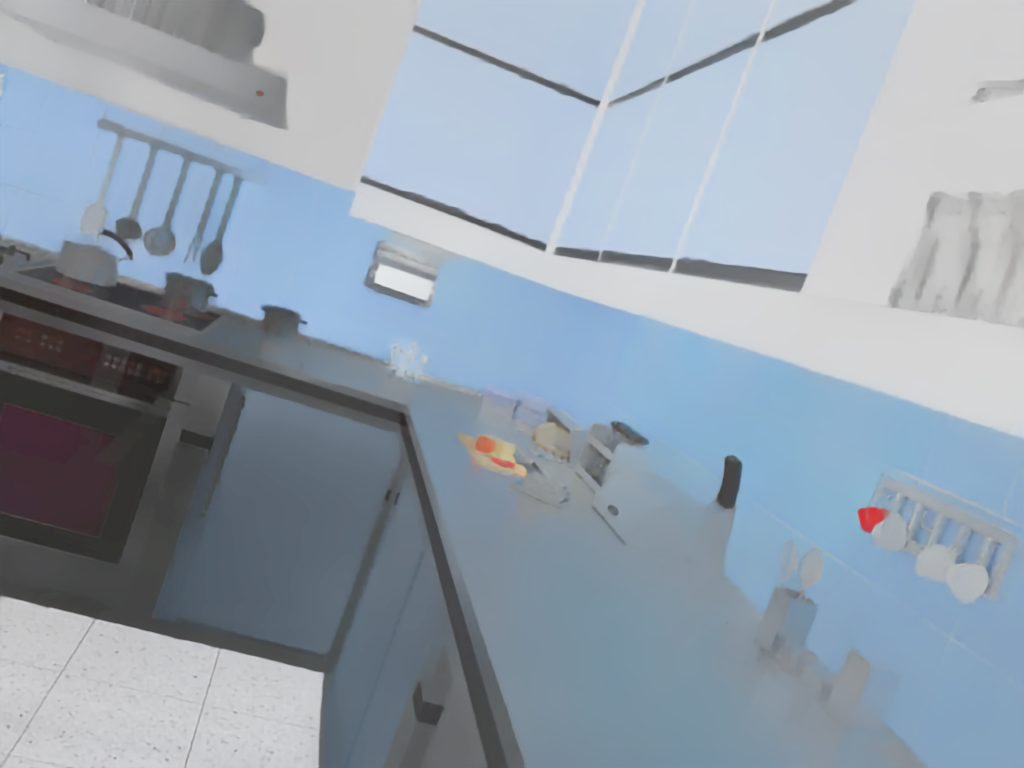}
                \put(6,79){\fontsize{8pt}{10pt}\selectfont \citet{careaga2024colorful}}
            \end{overpic}
            \hfill
            \begin{overpic}[width=0.16\linewidth]{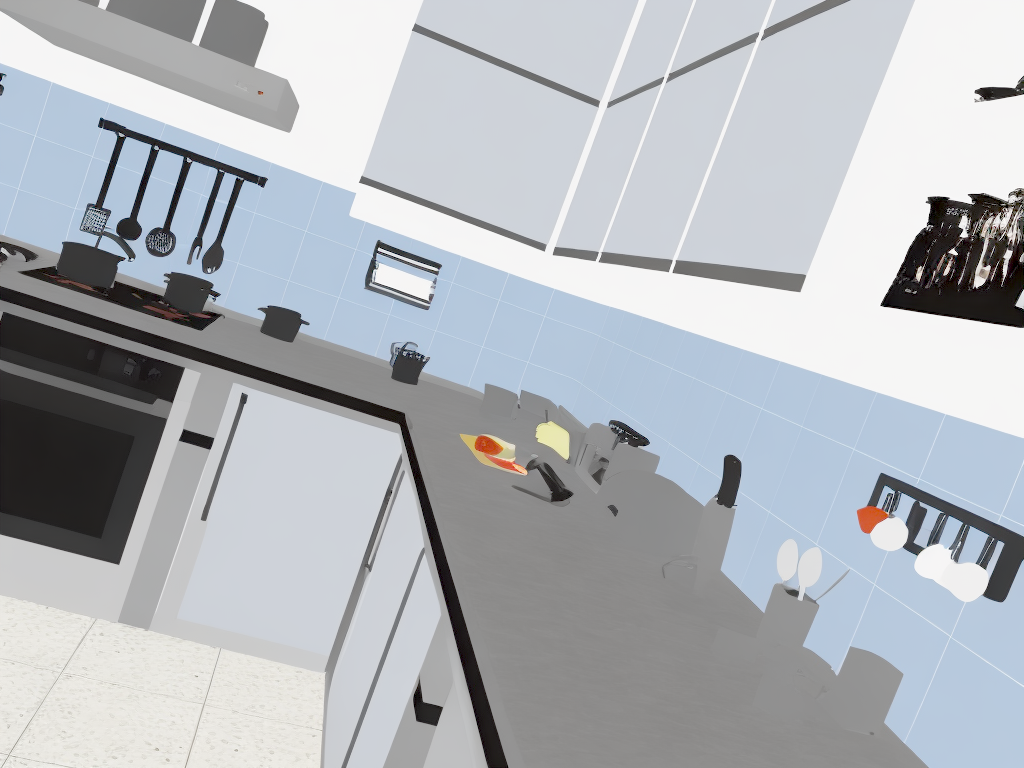}
                \put(19,79){\small RGB$\leftrightarrow$X~\cite{zeng2024rgb}}
            \end{overpic}
            \hfill
            \begin{overpic}[width=0.16\linewidth]{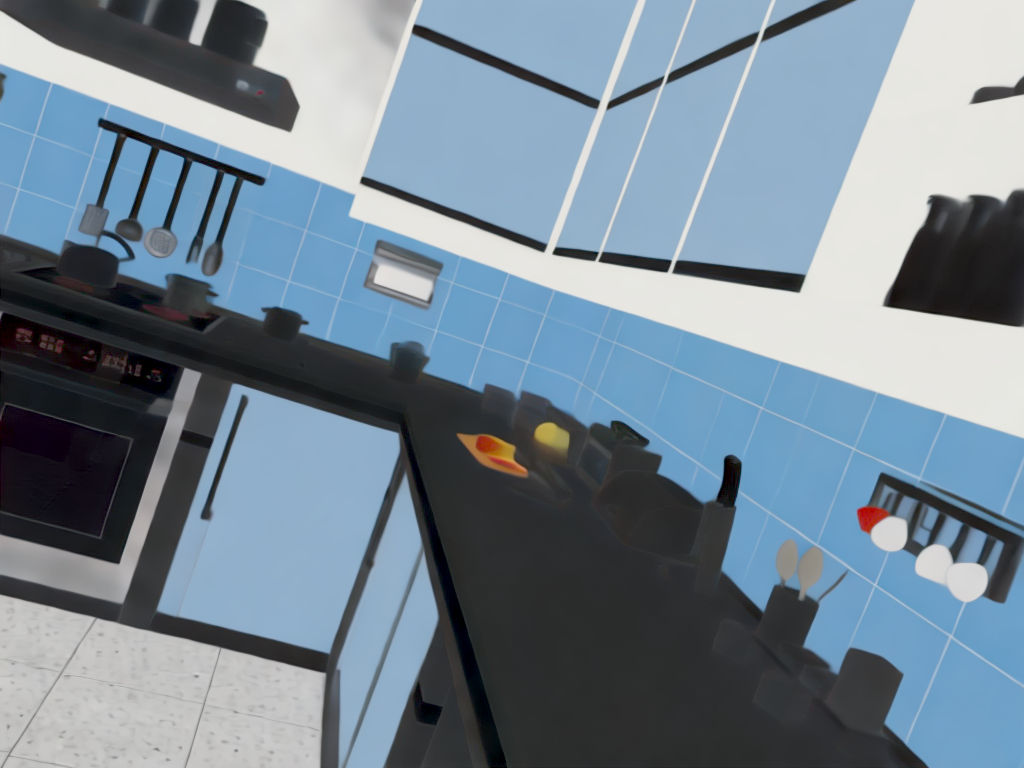}
                \put(37,79){\small\textbf{Ours}}
            \end{overpic}
            \hfill
            \begin{overpic}[width=0.16\linewidth]{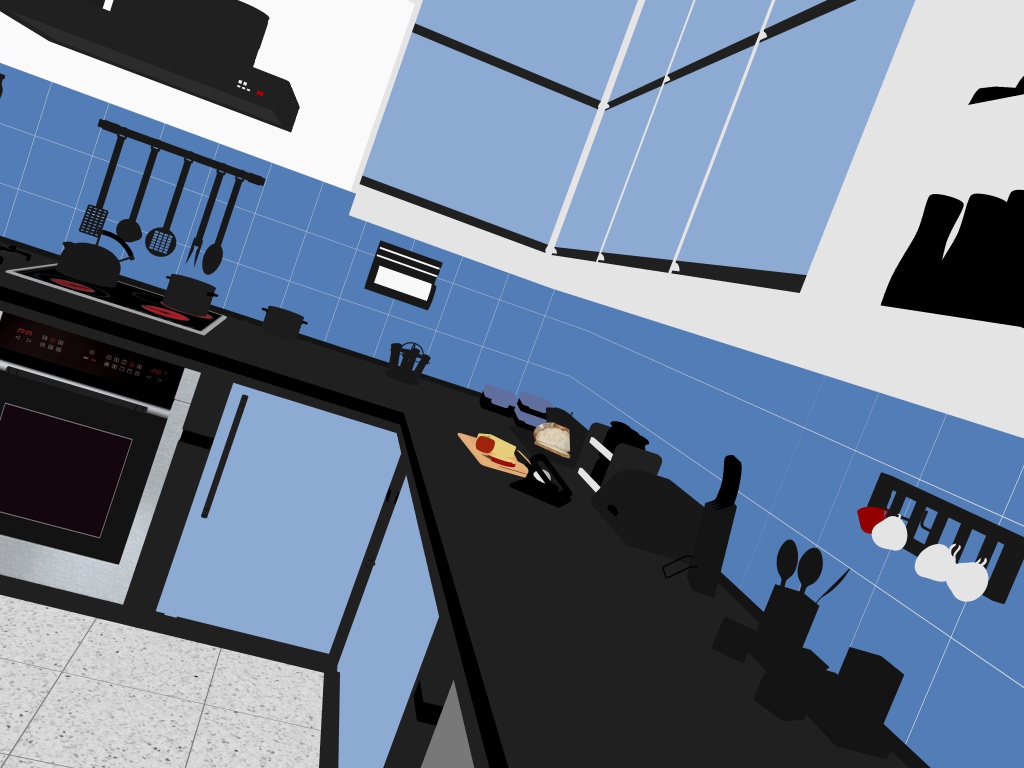}
                \put(20,79){\small Ground truth}
            \end{overpic}
        }
    \end{minipage}\par\smallskip
    \begin{minipage}{\linewidth}
        \begin{minipage}{\linewidth}
        \end{minipage}\par\smallskip
        \centering
        \subfloat[
        \textbf{Irradiance Estimation.}
        Comparing to RGB$\leftrightarrow$X, our method achieves comparable performance.
        ]{
            \begin{overpic}[width=0.16\linewidth]{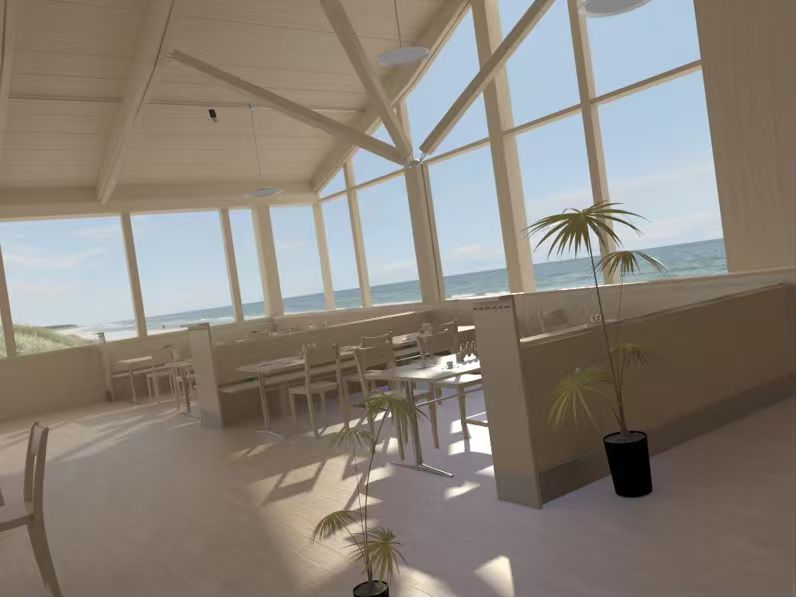}
                \put(22,79){\small Input image}
            \end{overpic}
            \hfill
            \begin{overpic}[width=0.16\linewidth]{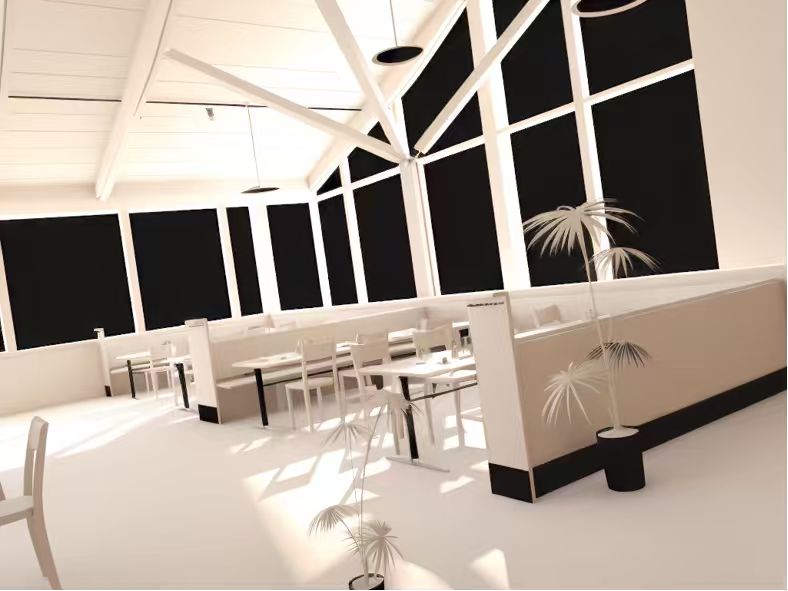}
                \put(28,79){\small RGB$\leftrightarrow$X}
            \end{overpic}
            \hfill
            \begin{overpic}[width=0.16\linewidth]{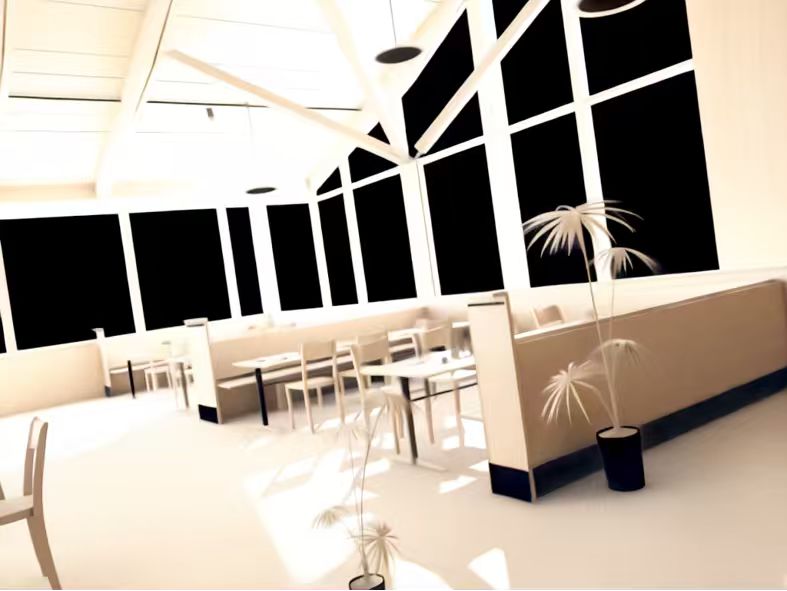}
                \put(37,79){\small \textbf{Ours}}
            \end{overpic}
            \hfill
            \begin{overpic}[width=0.16\linewidth]{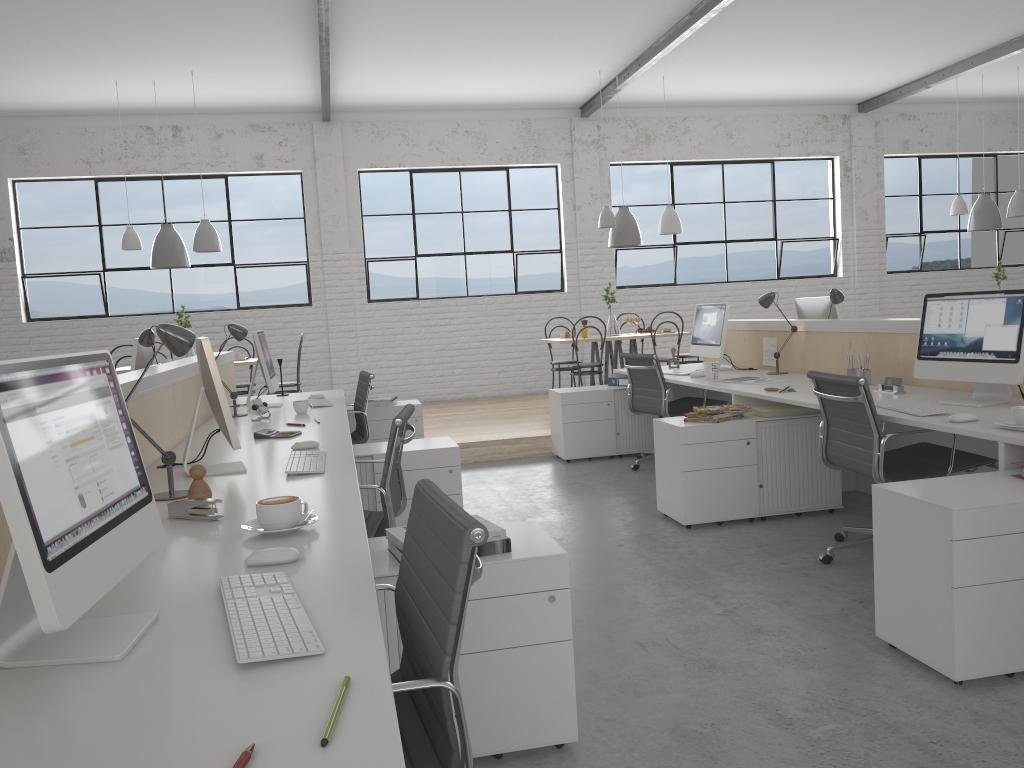}
                \put(22,79){\small Input image}
            \end{overpic}
            \hfill
            \begin{overpic}[width=0.16\linewidth]{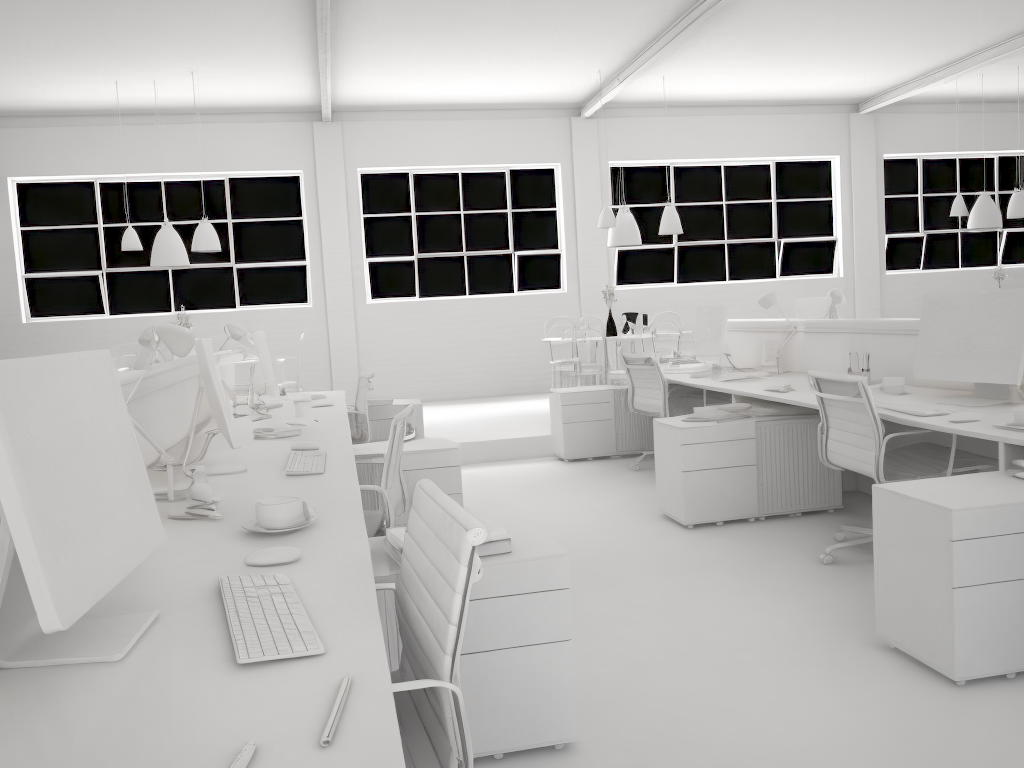}
                \put(28,79){\small RGB$\leftrightarrow$X}
            \end{overpic}
            \hfill
            \begin{overpic}[width=0.16\linewidth]{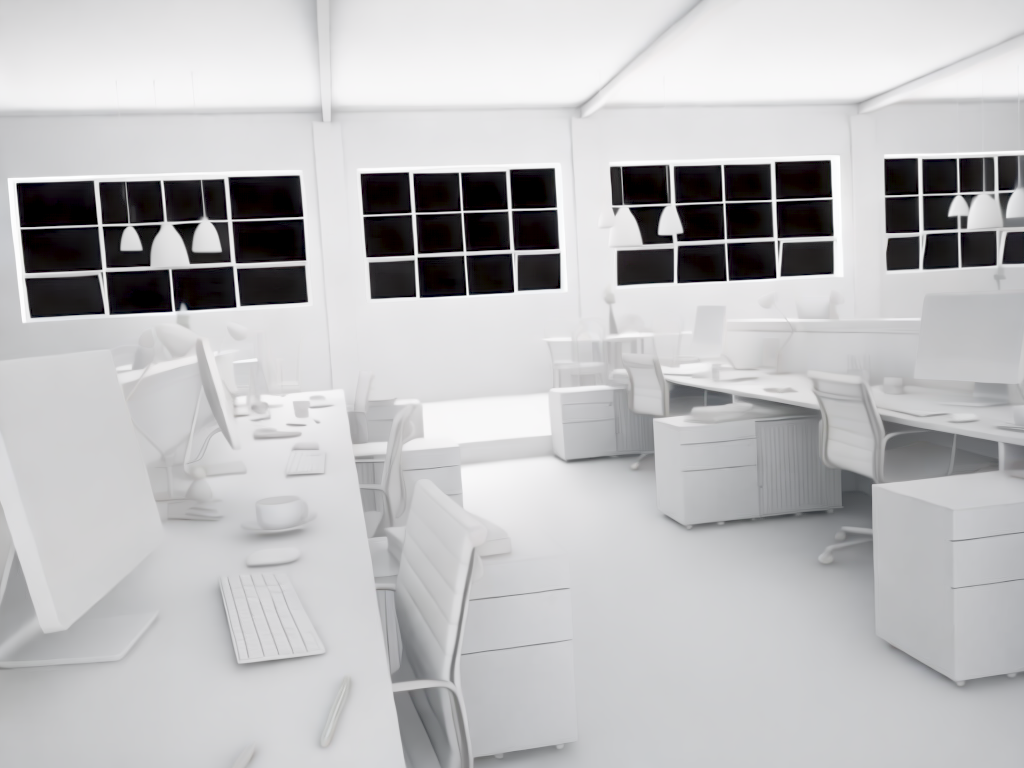}
                \put(37,79){\small \textbf{Ours}}
            \end{overpic}
        }
    \end{minipage}\par\smallskip
    \begin{minipage}{\linewidth}
        \begin{minipage}{\linewidth}
        \end{minipage}\par\medskip
        \begin{minipage}{0.475\linewidth}
            \centering
            \subfloat[
            \textbf{Roughness Estimation.} 
            Our method is more representative of the surface roughness of the image rather than confusing it with the material.
            ]{
            \begin{overpic}[width=0.332\linewidth]{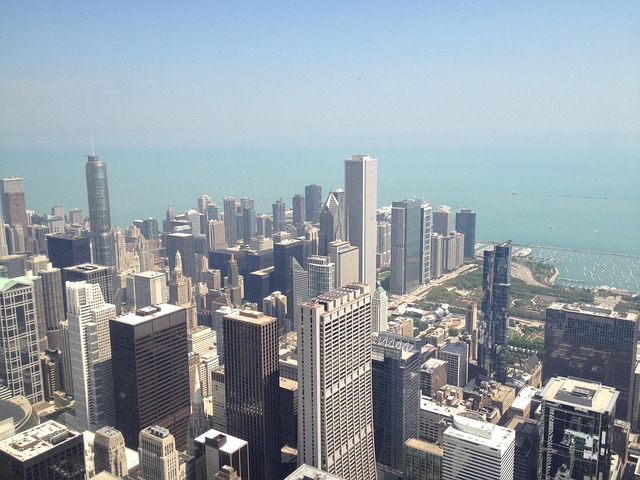}
            \put(22,79){\small Input image}
            \end{overpic}
            \hfill
            \begin{overpic}[width=0.332\linewidth]{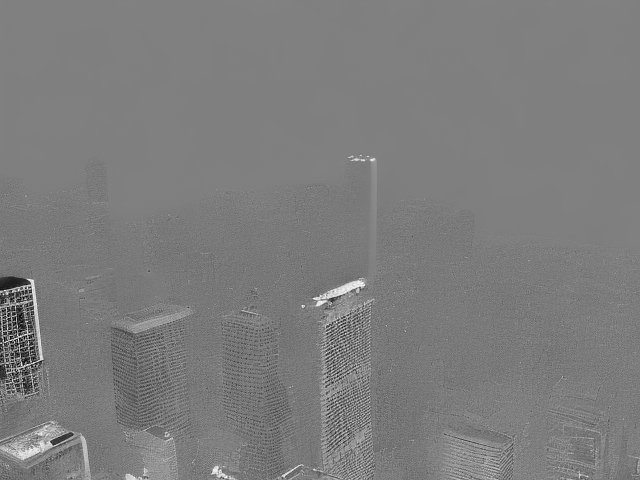}
                \put(28,79){\small RGB$\leftrightarrow$X}
            \end{overpic}
            \hfill
            \begin{overpic}[width=0.332\linewidth]{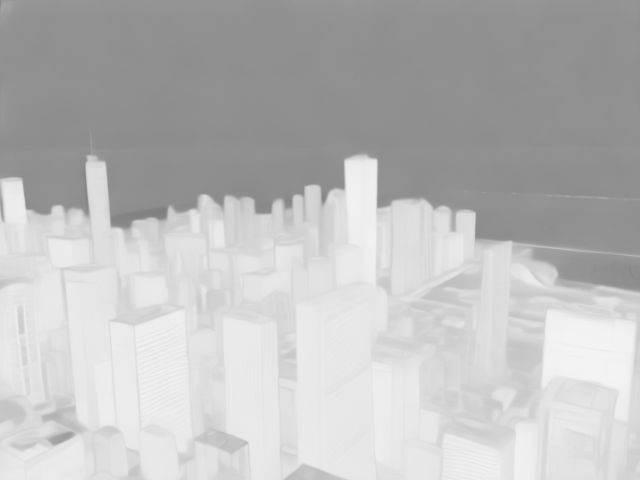}
                \put(37,79){\small \textbf{Ours}}
            \end{overpic}
            }
        \end{minipage}
        \hspace{9.5pt}
        \begin{minipage}{0.475\linewidth}
            \centering
            \subfloat[
            \textbf{Metallicity Estimation.} 
            Our method is more effective in accentuating the textural characteristics of diverse materials.
            ]{
            \begin{overpic}[width=0.332\linewidth]{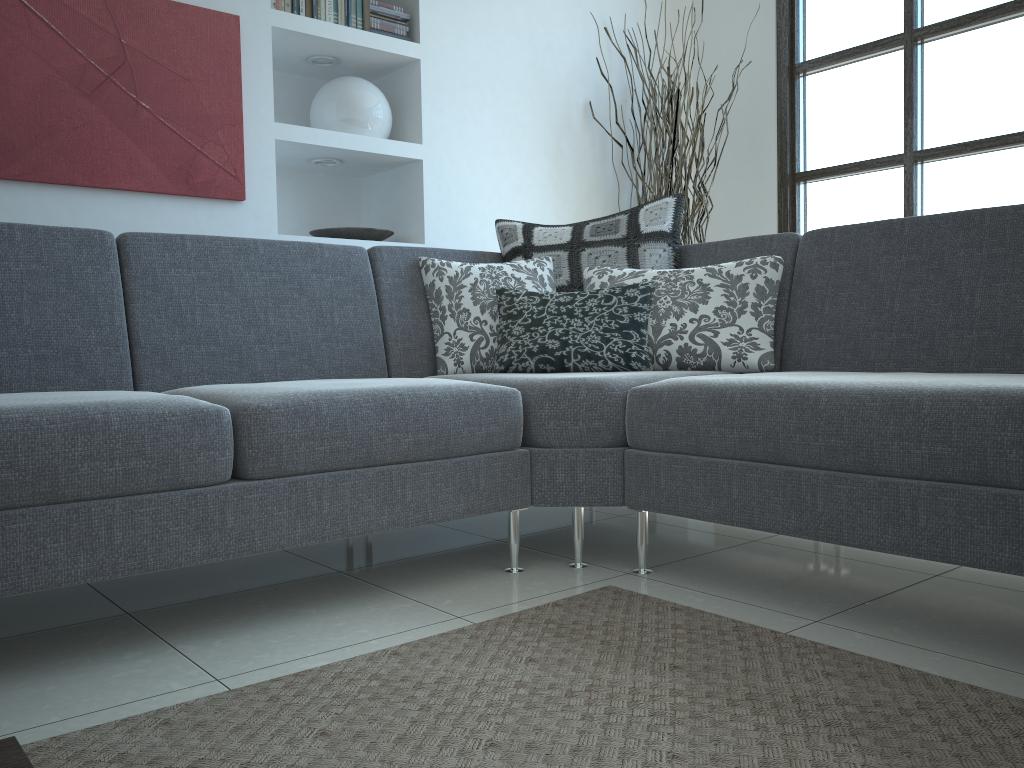}
            \put(22,79){\small Input image}
            \end{overpic}
            \hfill
            \begin{overpic}[width=0.332\linewidth]{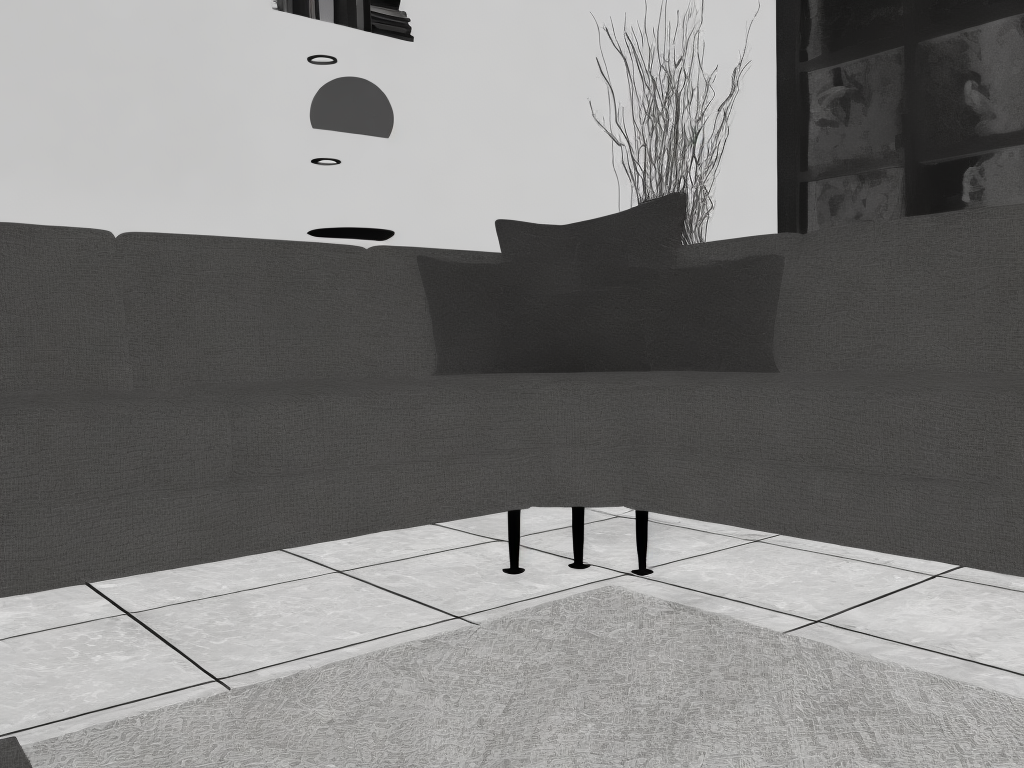}
                \put(28,79){\small RGB$\leftrightarrow$X}
            \end{overpic}
            \hfill
            \begin{overpic}[width=0.332\linewidth]{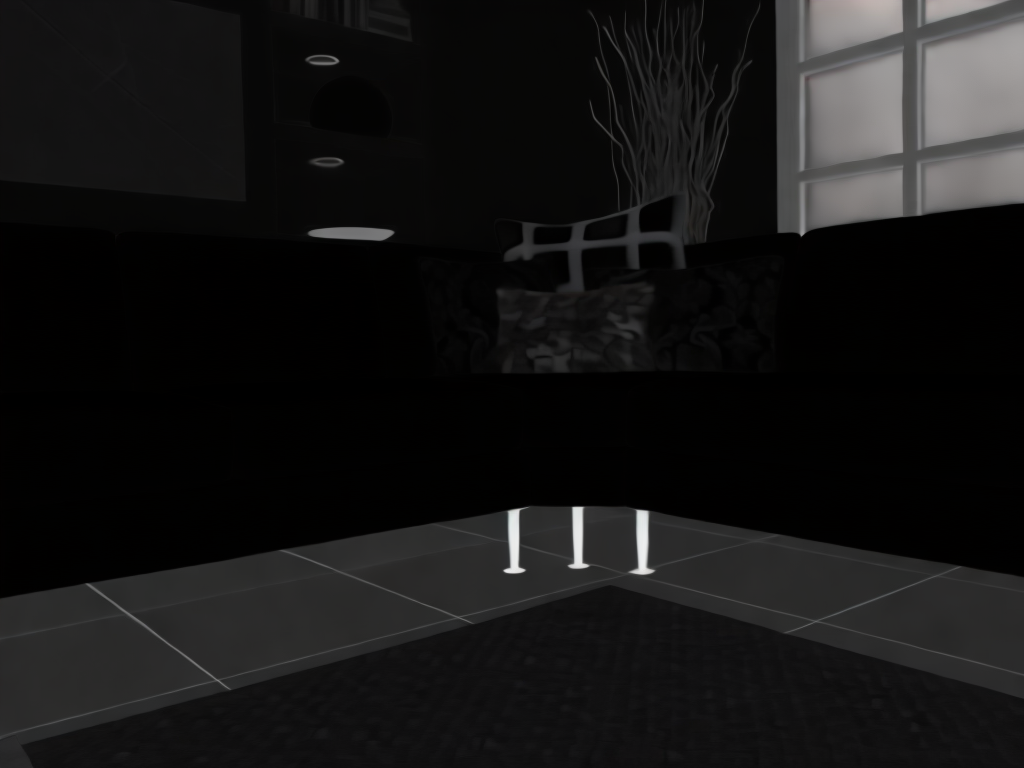}
                \put(37,79){\small \textbf{Ours}}
            \end{overpic}
            }
        \end{minipage}
    \end{minipage}\par\bigskip\medskip
    \begin{minipage}{\linewidth}
        \centering
        \subfloat[
        \textbf{Normal Estimation.}
        Our method outperforms the baseline methods in details and planar consistency.
        ]{
            \begin{overpic}[width=0.137\linewidth]{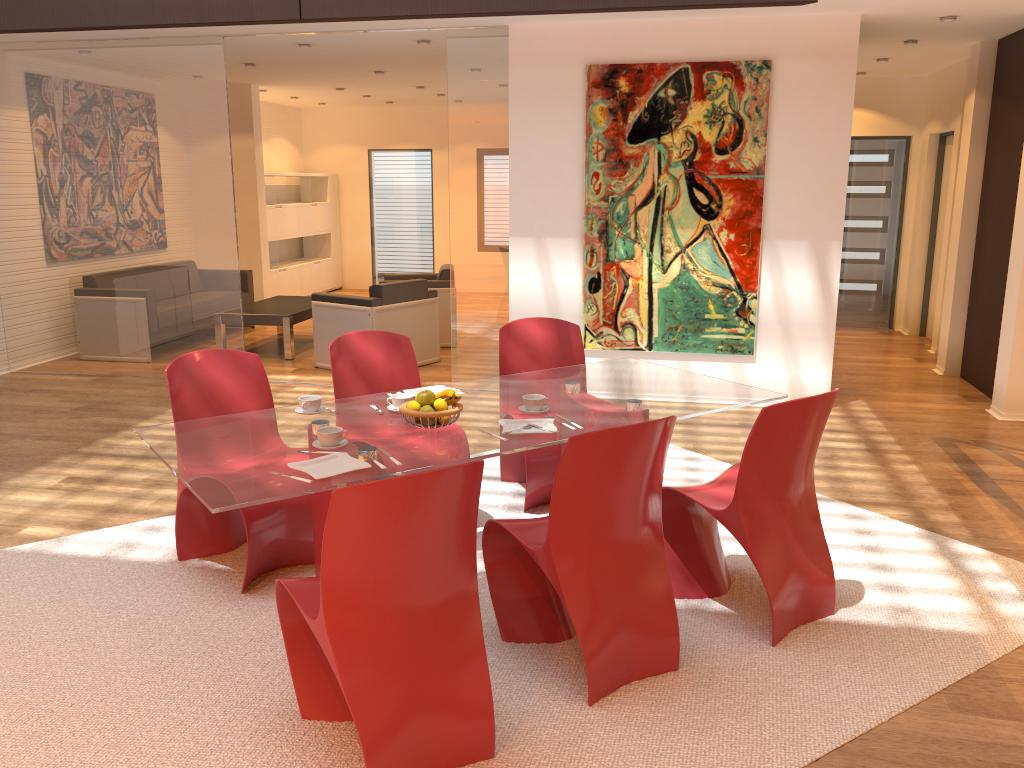}
                \put(20,79){\small Input image}
            \end{overpic}
            \begin{overpic}[width=0.137\linewidth]{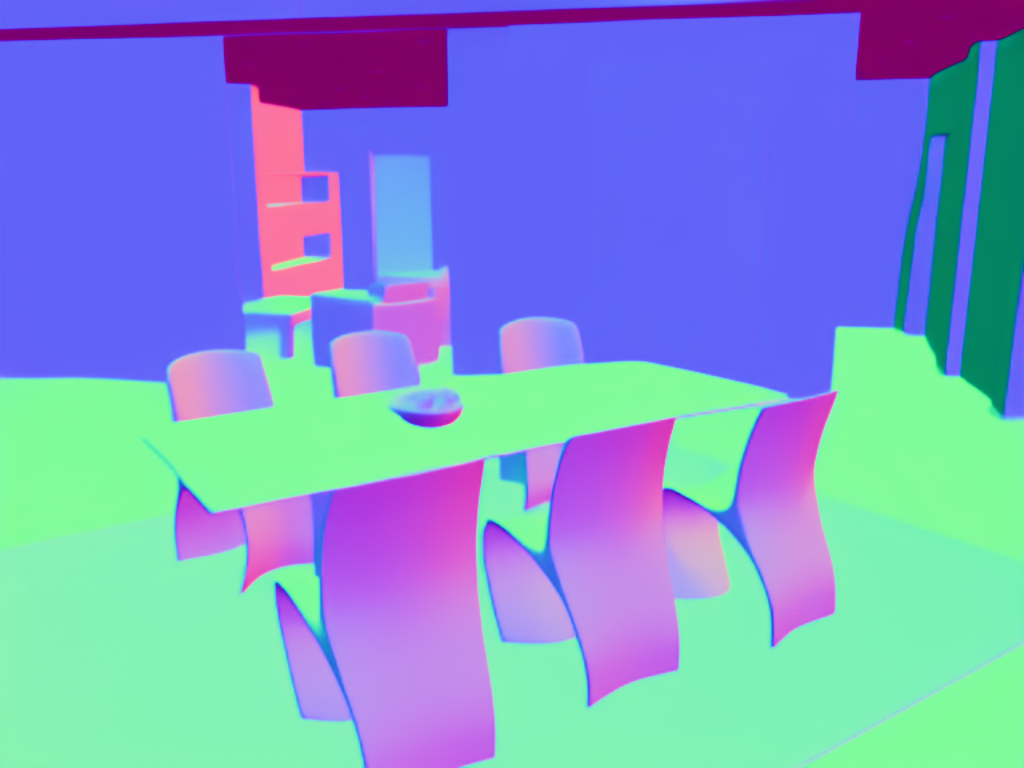}
                \put(25,79){\small E2E ~\cite{martingarcia2024diffusione2eft}}
            \end{overpic}
            \begin{overpic}[width=0.137\linewidth]{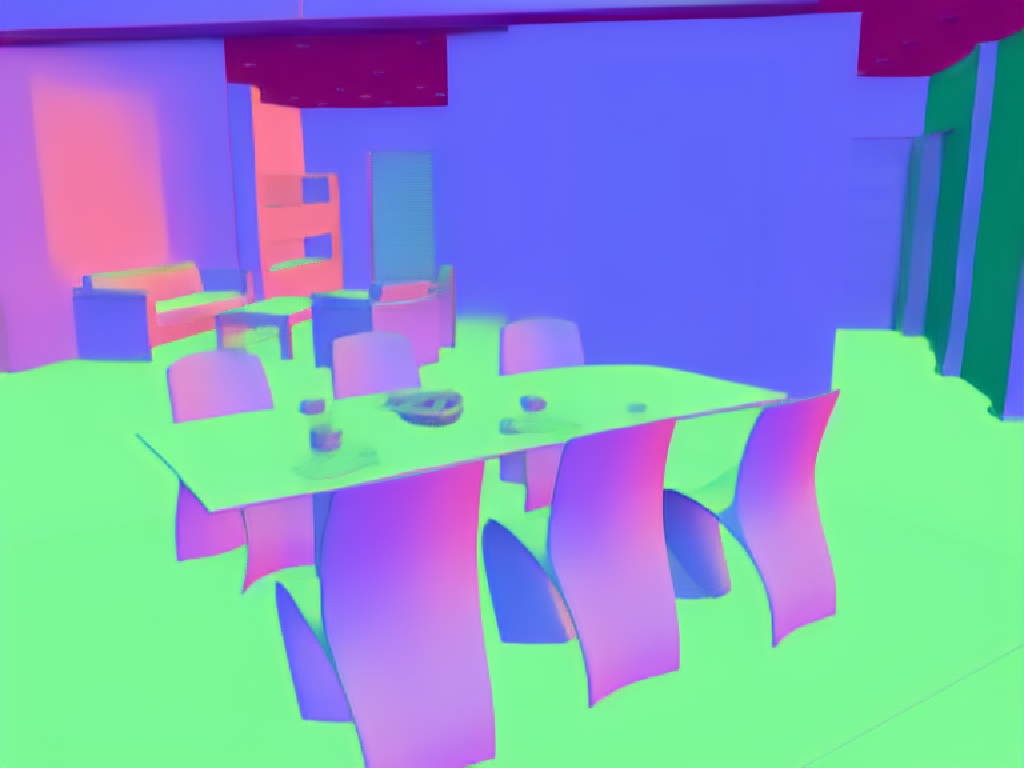}
                \put(25,79){\small Lotus~\cite{he2024lotus}}
            \end{overpic}
            \begin{overpic}[width=0.137\linewidth]{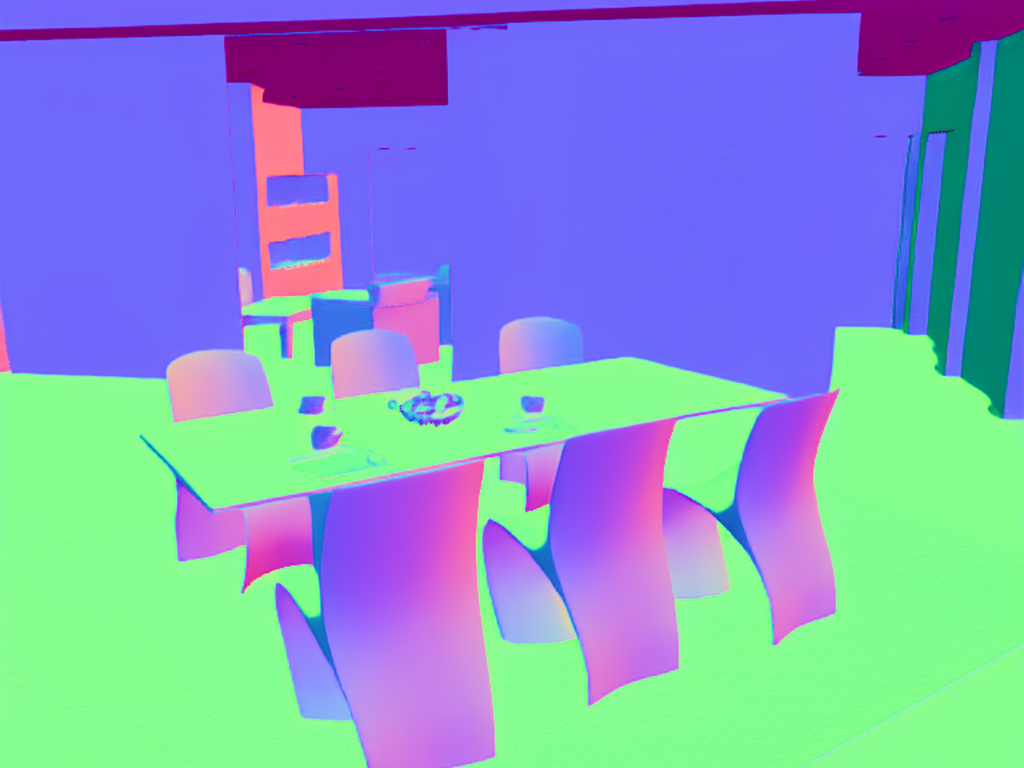}
                \put(5,79){\fontsize{8pt}{10pt}\selectfont StableNormal~\cite{ye2024stablenormal}}
            \end{overpic}
            \begin{overpic}[width=0.137\linewidth]{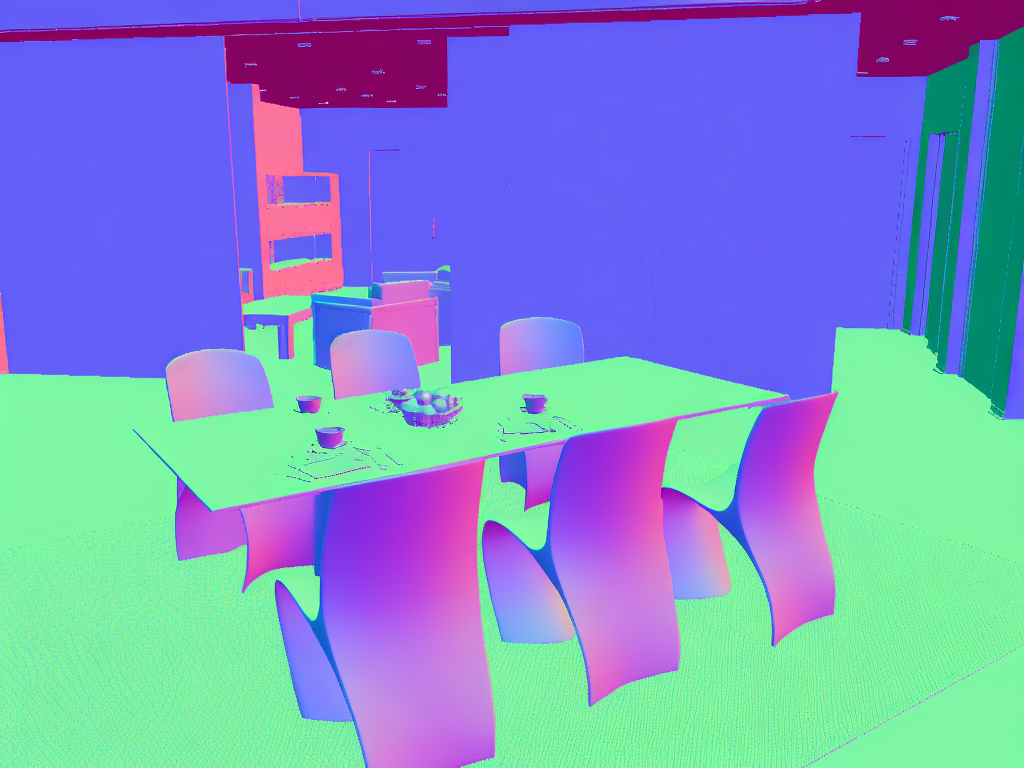}
                \put(25,79){\small RGB$\leftrightarrow$X}
            \end{overpic}
            \begin{overpic}[width=0.137\linewidth]{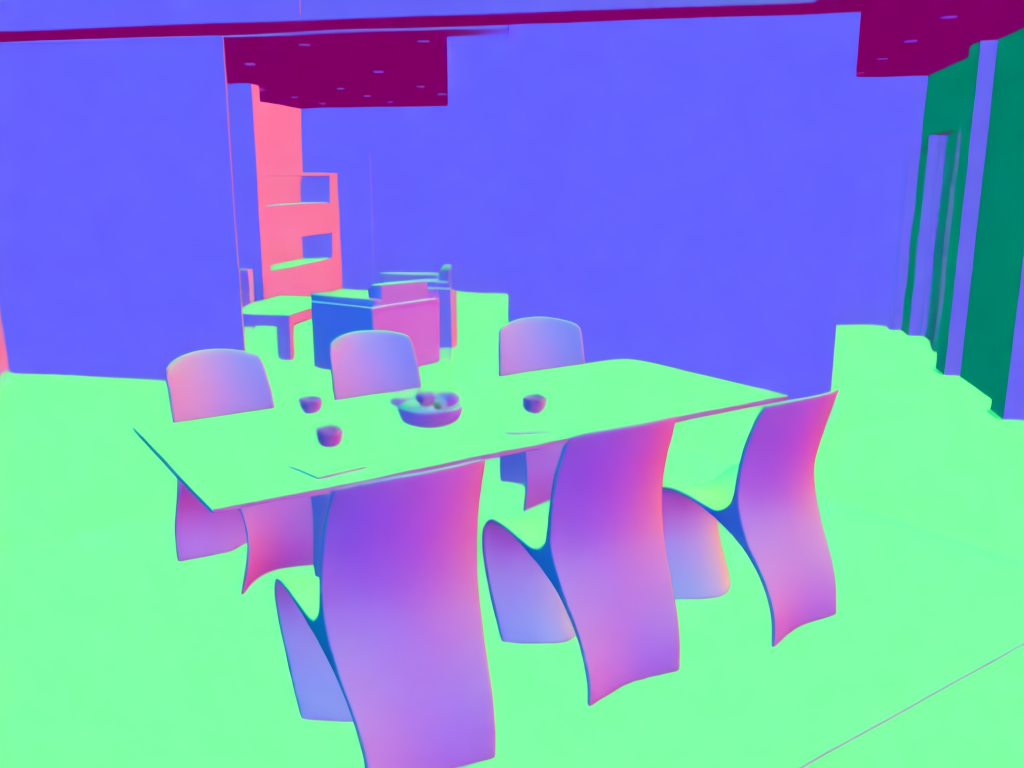}
                \put(36,79){\small \textbf{Ours}}
            \end{overpic}
            \begin{overpic}[width=0.137\linewidth]{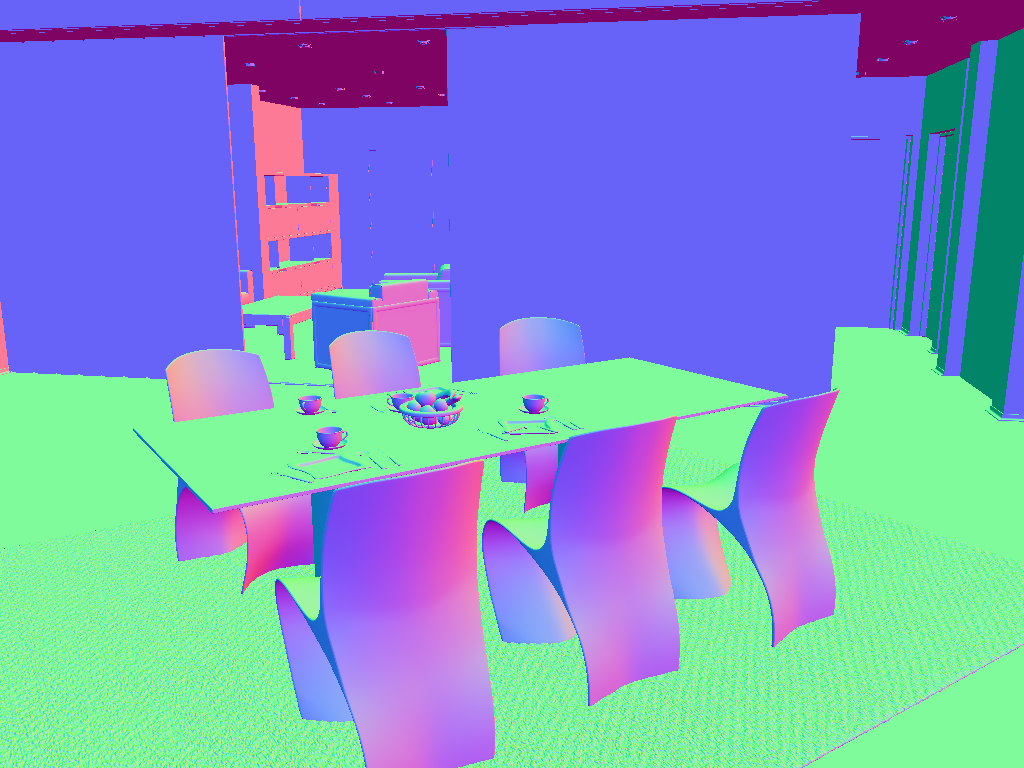}
                \put(14,79){\small Ground truth}
            \end{overpic}
        }
    \end{minipage}\par
\end{minipage}
\caption{\textbf{Comprehensive Visual Comparison between Baseline Models and our \model ~on Diverse Inverse Rendering Tasks.}}
    \label{fig:rgb2x_comp}
\end{figure*}

\begin{table}[!htbp]
    \centering
    \caption{Comparison between Our \model ~and RGB$\leftrightarrow$X in Forward Rendering.}
    \vspace{-0.5em}
    \resizebox{\columnwidth}{!}{
    \begin{tabular}{@{}l|cc|cc|cc}
        \toprule
        \multirow{2}*{\textbf{Method}}  & \multicolumn{2}{c|}{\textbf{Hypersim}} & \multicolumn{2}{c|}{\textbf{MatrixCity}} & \multicolumn{2}{c}{\textbf{InteriorVerse}} \\
        \cmidrule(lr){2-3} \cmidrule(lr){4-5} \cmidrule(lr){6-7}
        
        & \textbf{PSNR$\uparrow$} & \textbf{LPIPS$\downarrow$} & \textbf{PSNR$\uparrow$} & \textbf{LPIPS$\downarrow$} & \textbf{PSNR$\uparrow$} & \textbf{LPIPS$\downarrow$}  \\
        \midrule
        RGB$\leftrightarrow$X & 16.37 & \textbf{0.20} & 9.24 & 0.30 & 13.70 & 0.33 \\
        \textbf{Ours} & \textbf{18.09} & 0.25 & \textbf{21.57} & \textbf{0.18} &  \textbf{15.79} & \textbf{0.28} \\
        \bottomrule
    \end{tabular}
    }
    \label{tab:forward_rendering}
\end{table}
\begin{figure}
  \centering
    \includegraphics[width=1.0\linewidth]{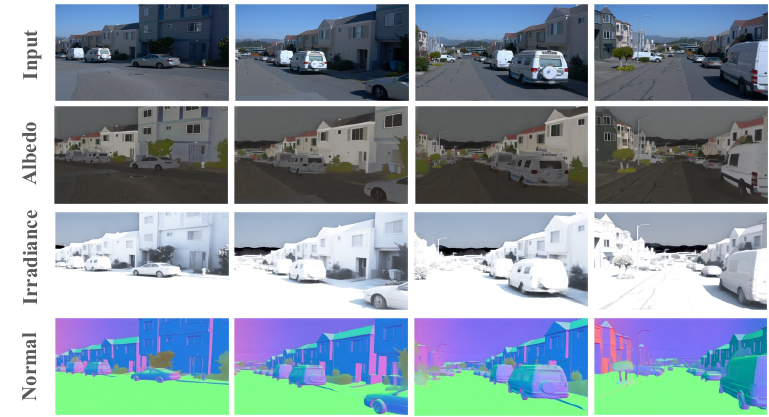}
    \vspace{-1.5em}
    \caption{\textbf{Examples of Video Inference.} Our model demonstrates the ability to process real-world scenarios.}
    \label{fig:video_qualitative_results}
\end{figure} 
\begin{figure*}[!htbp]
  \centering
   \includegraphics[width=1.\linewidth]{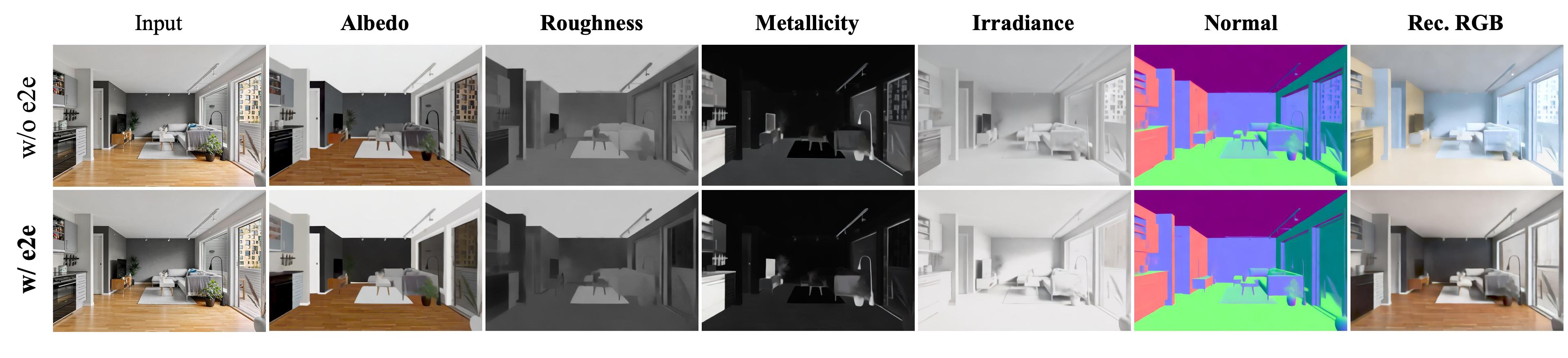}
   \vspace{-2em}
   \caption{\textbf{Ablation Study on Cycle Training with or w/o e2e Loss.} Methods incorporating e2e loss can better understand lighting conditions and provide more continuous estimation. We can observe that the colors in the restored images are also more accurate and faithful.}
   \label{fig:ablation_e2e}
\end{figure*}
\begin{figure}
  \centering
   \includegraphics[width=1.\linewidth]{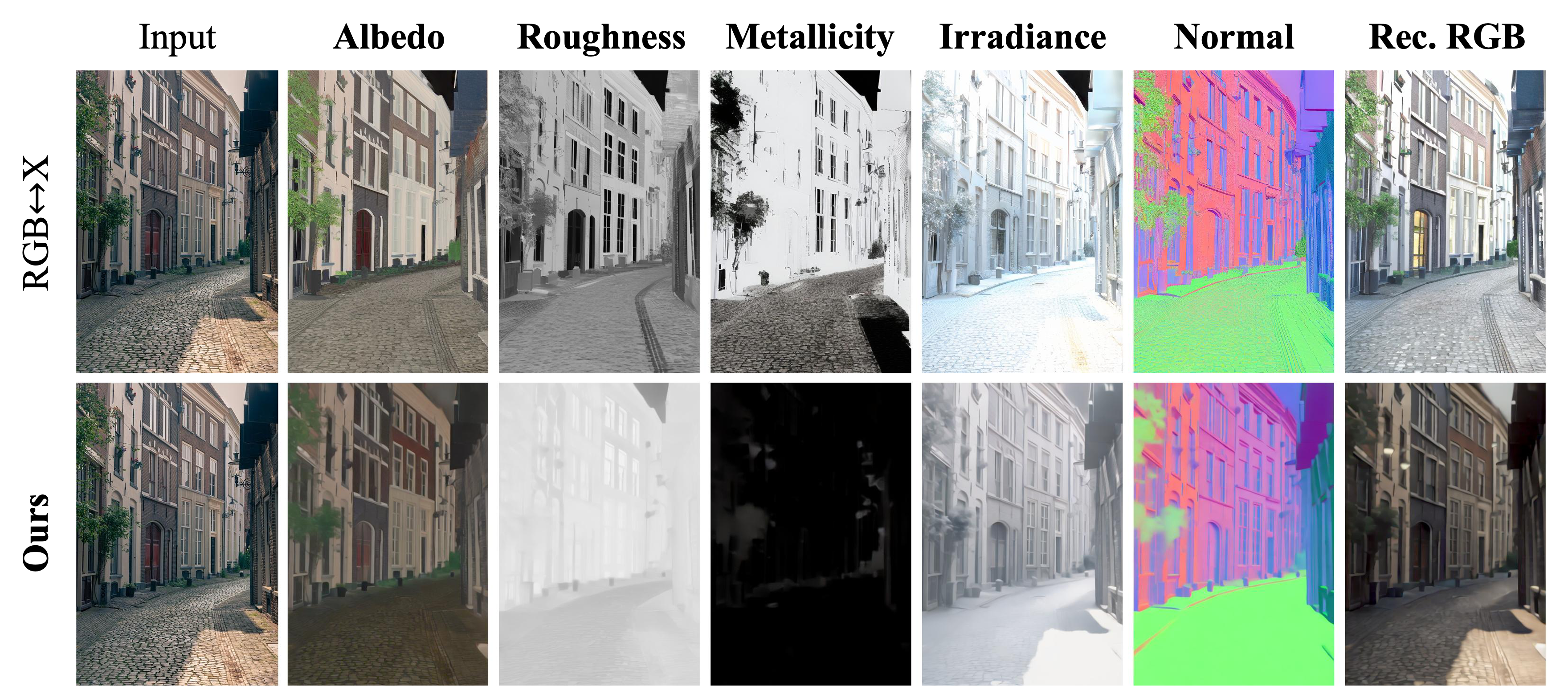}
   \caption{\textbf{Visual Comparison between RGB$\leftrightarrow$X and ours on Wild Data.} Our method demonstrates superior performance in terms of material understanding, lighting comprehension, rendering consistency.}
   \label{fig:wild_data_comp}
\end{figure}
\begin{figure}[!htbp]
  \centering
  \setlength{\tabcolsep}{1pt}
  \resizebox{\columnwidth}{!}{%
    \begin{tabular}{ccccc}
      Input & \textbf{Irr. w/ Cycle} & Irr. w/o Cycle & \textbf{Rec. w/ Cycle} & Rec. w/o Cycle\\
      \includegraphics[width=.3\linewidth]{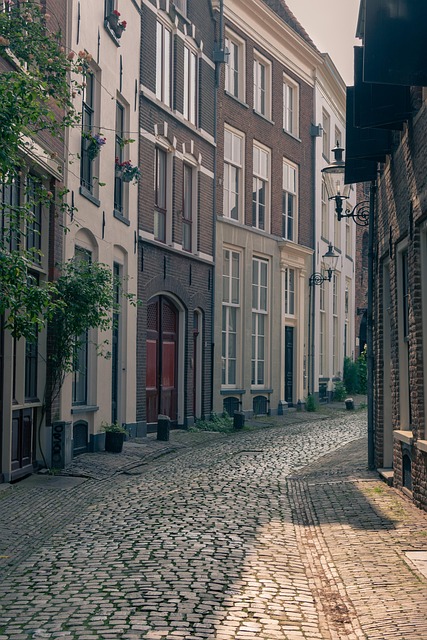} & 
      \includegraphics[width=.3\linewidth]{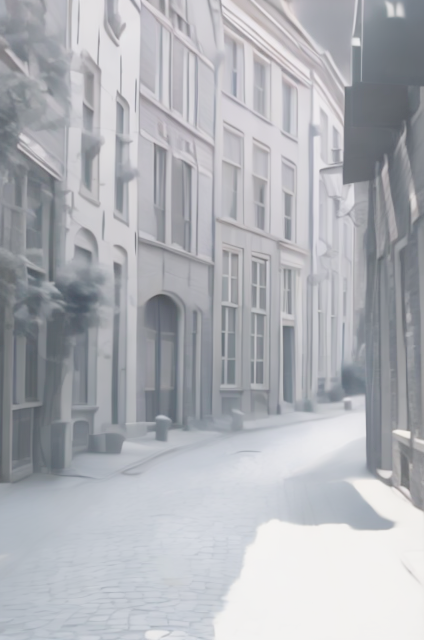} & 
      \includegraphics[width=.3\linewidth]{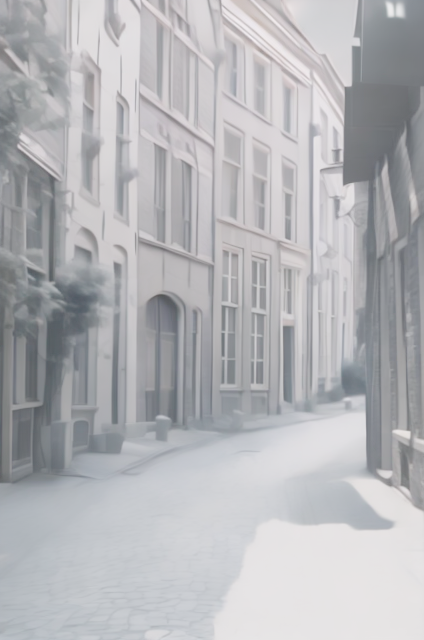} & 
      \includegraphics[width=.3\linewidth]{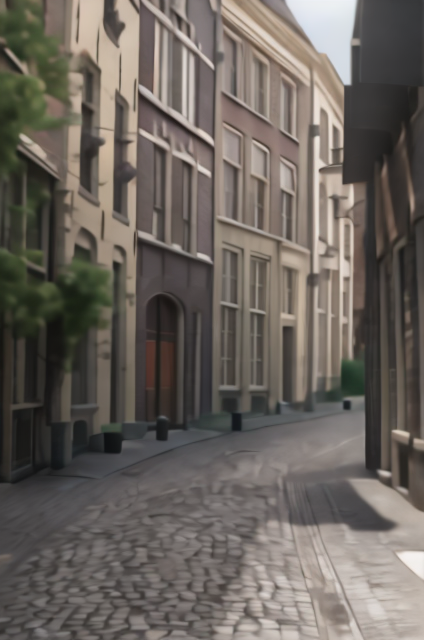} &
      \includegraphics[width=.3\linewidth]{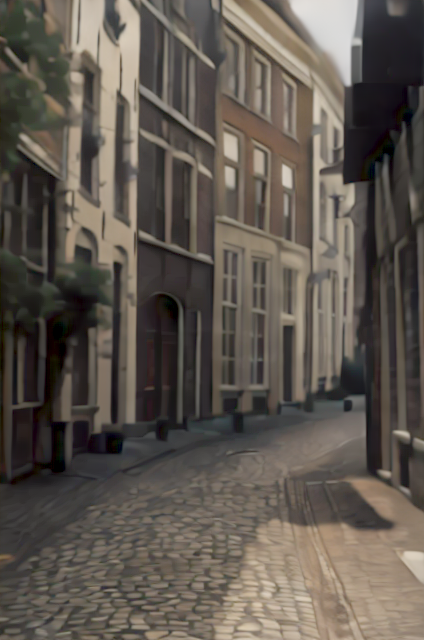} 
    \end{tabular}%
  }
  \vspace{-0.5em}
  \caption{\textbf{Ablation Study on Performance with or without Cycle Training.} With cycle training, the irradiance will be more sharp in details and the color of reconstruction is more consistant with the input.}
  \label{fig:cycle}
\end{figure}
\begin{figure}
  \centering
   \includegraphics[width=1.\linewidth]{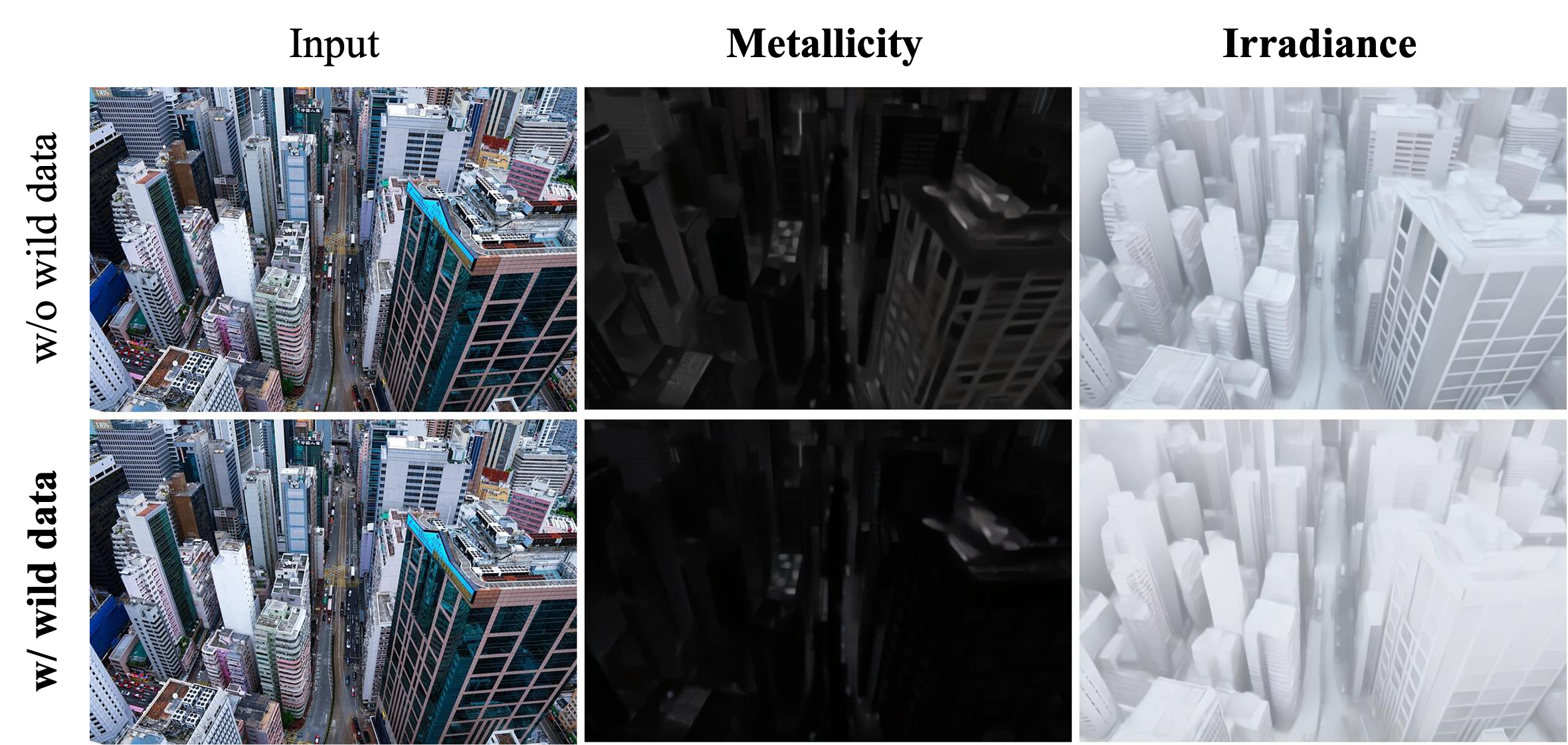}
   \caption{\textbf{Ablation Study on Cycle Training with or without Wild Data.} Training on wild data helps improve the understanding of overhead lighting and surface materials.}
   \label{fig:ablation_wild}
\end{figure}

\subsection{Setup}
Given that the utilization of a combination of heterogeneous datasets, and the efficacy of cycle training in mitigating the impact of imperfections in certain data, it was determined that all modalities would be utilized in the training process.

\paragraph{Evaluation Datasets.} For inverse rendering, we follow the setting same as RGB$\leftrightarrow$X~\cite{zeng2024rgb} and utilized the Hypersim~\cite{roberts2021hypersim} test set to evaluate albedo, normal and irradiance estimation. 
Furthermore, to ensure the reliable evaluation for both indoor and outdoor scenes, the test sets of InteriorVerse~\cite{zhu2022learning} and MatrixCity~\cite{li2023matrixcity} were employed for albedo, normal roughness and metallicity estimation as well. 
As for the forward rendering evaluation, the same test set of the three datasets mentioned above was applied. The intrinsic maps utilized as input varied by dataset, Hypersim used albedo, normal, and irradiance maps, while MatrixCity and InteriorVerse employed albedo, normal, roughness, and metallicity maps.

\paragraph{Baseline Methods.}
Since our model is finetuned based on the RGB$\leftrightarrow$X~\cite{zeng2024rgb} pretrain weights, we compared with it in all intrinsic map estimation tasks. For normal map estimation, we compare with state-of-the-art methods including the model proposed in \citet{zhu2022learning}, StableNormal~\cite{ye2024stablenormal}, E2E~\cite{martingarcia2024diffusione2eft}, and Lotus~\cite{he2024lotus}.
For albedo estimation, we include comparisons with models proposed by \citet{careaga2024colorful} and \citet{kocsis2023intrinsic}.
And for roughness and metallicity estimation, we compare our results with models proposed by \citet{kocsis2023intrinsic} and \citet{zhu2022learning}.

\paragraph{Metrics.}
The evaluation of inverse rendering involved the utilization of PSNR and LPIPS metrics to assess all intrinsic maps. For albedo evaluation, scale-invariant root mean squared error (RMSE) and SSIM metrics were also incorporated. The evaluation of normal incorporated mean angular errors (Mean) and the percentage of pixels with angular errors below 11.25° thresholds. Forward rendering evaluation primarily employed PSNR, LPIPS, and SSIM metrics to assess the reconstruction quality.

\subsection{Inverse Rendering Results}
\subsubsection{Quantitative Results}
In the context of albedo and normal estimation, our approach demonstrates superior performance in nearly all metrics across three distinct datasets, with only a limited number of metrics exhibiting a second-place ranking. 
It is noteworthy that our method requires only a single step of inference, which further underscores its ability to produce high-quality results while significantly decrease inference time. 
The quantitative results are presented in detail in Tab.~\ref{albedo prediction} and ~\ref{normal prediction}. 

\subsubsection{Qualitative Results}
We conduct qualitative comparisons of our model's inverse rendering capabilities across diverse indoor and outdoor scenes, against multiple baseline models, including RGB$\leftrightarrow$X~\cite{zeng2024rgb}, as shown in \cref{fig:indoor_syn}. Due to space limitations, comparisons in outdoor scenes can be seen in the supplementary materials Fig.~7. All test scenes are explicitly excluded from our training dataset. 

In terms of albedo prediction, our method achieves performance parity with RGB$\leftrightarrow$X~\cite{zeng2024rgb} in indoor environments while demonstrating clear advantages over other baseline models. In outdoor scenarios, our approach consistently outperforms RGB$\leftrightarrow$X~\cite{zeng2024rgb}, particularly excelling at preserving fine details in the generation of albedo for distant objects. Moreover, our method exhibits superior ability in handling specular reflections and material properties, whereas RGB$\leftrightarrow$X~\cite{zeng2024rgb} tends to conflate reflection information into albedo estimation, resulting in discrepancies from the true albedo values. 

Regarding normal estimation, our method achieves performance comparable to SOTA models in indoor scenes, while in outdoor scenarios, we demonstrate more refined detail in normal prediction for smaller objects within larger scenes. Furthermore, we observe that RGB$\leftrightarrow$X~\cite{zeng2024rgb} tends to overestimate surface roughness in ground normal prediction, generating dense, discontinuous normals even for flat surfaces with rich color variations.

In the estimation of roughness and metallic properties, despite minor local deviations, our method demonstrates superior performance across a wider spectrum of scenarios. For indoor environments, particularly in the prediction of smooth surfaces such as tables and cabinets, our approach consistently outperforms RGB$\leftrightarrow$X~\cite{zeng2024rgb}, which exhibits material interpretation inconsistencies. In outdoor environments, especially in the analysis of tall and distant architectural structures, RGB$\leftrightarrow$X~\cite{zeng2024rgb} shows substantial estimation biases, while our method maintains more consistent and reliable predictions.

Our method for irradiance understanding matches the performance of RGB$\leftrightarrow$X~\cite{zeng2024rgb} indoors and proves more reliable in outdoor scenarios, particularly in capturing lighting on skyscraper surfaces and windows. Since our model was trained to estimate irradiance exclusively on indoor scenes in Hypersim, these results validate that our cycle-based approach successfully generalizes its understanding to new environments.

\subsection{Forward Rendering Results}
\subsubsection{Quantitative Results}
In the forward rendering comparison, our method outperforms RGB$\leftrightarrow$X~\cite{zeng2024rgb} across most metrics, with particularly significant advantages in performance in MatrixCity.
The quantitative results are presented in detail in Tab.~\ref{tab:forward_rendering}. 

\subsubsection{Qualitative Results}
Compared to manually rendering individual image channels, which is often time and resource-intensive, we opt for an approach that performs both inverse rendering and forward rendering from a single RGB input. This method not only helps us intuitively understand the forward rendering capabilities but also enables us to observe the differences between the generated RGB and the initial input RGB, allowing us to evaluate the continuity of the process.

Our experimental results demonstrate superior performance over RGB$\leftrightarrow$X~\cite{zeng2024rgb} in both forward rendering fidelity and consistency between input and synthesized RGB images. This advantage is particularly evident in outdoor scenes with challenging illumination conditions, where our method successfully reconstructs the original lighting distributions. Moreover, our approach exhibits enhanced capability in color fidelity preservation for distant objects compared to RGB$\leftrightarrow$X~\cite{zeng2024rgb}. Some results can be found at Fig~\ref{fig:headline}.

Fig.~\ref{fig:video_qualitative_results} showcases our model’s ability to perform video inference in real-world scenes, preserving spatial details and ensuring temporal consistency under complex lighting and material variations. 
\subsection{Ablation Study on Cycle Training}

In \cref{fig:wild_data_comp}, we demonstrate advantages in both inverse rendering and forward rendering compared to RGB$\leftrightarrow$X. In inverse rendering, our \model~ shows better understanding of object materials such as building textures. In forward rendering, our method is significantly better at the recovery of lighting effects.

\paragraph{Effects of Cycle Training.}
We conducted a comparative analysis between cycle training and the combined inverse rendering and forward rendering approach for RGB-to-RGB generation. Our findings reveal that the cycle training paradigm achieves more faithful reproduction of lighting contrasts in outdoor scenes, closely matching the original image characteristics. Details can be found in Fig.~\ref{fig:cycle}.

\paragraph{Effects of Wild Data Finetuning.}
\cref{fig:ablation_wild} showcases the effectiveness of training with wild data. For tall buildings, we can clearly see that training with wild data produces more realistic irradiance, while metallicity is more continuous, demonstrating better understanding of surface properties.

\paragraph{Effects of e2e Loss.}
As shown in \cref{fig:ablation_e2e}, we evaluated whether to use e2e loss in cycle training. We can observe that with e2e loss, there is better continuity in metallicity and irradiance predictions, and the results are more faithful to the actual physical properties and materials.

\section{Discussion}
\model~explores single-step diffusion models for inverse and forward rendering, demonstrating superior performance compared to state-of-the-art methods across both indoor and outdoor scenes, as well as in image and video domains.

\paragraph{Limitations and future work.}
We identify training data quality and quantity as the primary bottleneck for neural rendering models. Current public datasets~\cite{zhu2022learning,roberts2021hypersim} often contain unreliable intrinsic maps and typically lack accurate lighting information, limiting the model's potential. To better support future image editing applications such as relighting and object insertion, we plan to curate a large-scale, diverse, high-quality synthetic dataset with full control over all components of 3D scenes using procedural generation techniques. This dedicated dataset would enable scalable training of neural rendering models and potentially resolve the current limitations in output fidelity.
\section{Acknowledgment}
We gratefully acknowledge UFIT Research Computing at the University of Florida and Stony Brook Research Computing and Cyberinfrastructure and the Institute for Advanced Computational Science at Stony Brook University for providing the computational resources and support that contributed to the research presented in this publication.
\clearpage
{
    \small
    \bibliographystyle{ieeenat_fullname}
    \bibliography{main}

@inproceedings{zeng2024rgb,
author = {Zeng, Zheng and Deschaintre, Valentin and Georgiev, Iliyan and Hold-Geoffroy, Yannick and Hu, Yiwei and Luan, Fujun and Yan, Ling-Qi and Ha\v{s}an, Milo\v{s}},
title = {RGB{$\leftrightarrow$}X: Image decomposition and synthesis using material- and lighting-aware diffusion models},
year = {2024},
isbn = {9798400705250},
publisher = {Association for Computing Machinery},
address = {New York, NY, USA},
url = {https://doi.org/10.1145/3641519.3657445},
doi = {10.1145/3641519.3657445},
booktitle = {ACM SIGGRAPH 2024 Conference Papers},
articleno = {75},
numpages = {11},
location = {Denver, CO, USA},
series = {SIGGRAPH '24}
}

@article{barrow1978recovering,
  title={Recovering intrinsic scene characteristics},
  author={Barrow, Harry and Tenenbaum, J and Hanson, A and Riseman, E},
  journal={Comput. vis. syst},
  volume={2},
  number={3-26},
  pages={2},
  year={1978}
}

@article{kocsis2023intrinsic,
  title={Intrinsic Image Diffusion for Single-view Material Estimation},
  author={Kocsis, Peter and Sitzmann, Vincent and Nie{\ss}ner, Matthias},
  journal={arXiv preprint arXiv:2312.12274},
  year={2023}
}

@InProceedings{martingarcia2024diffusione2eft,
  title     = {Fine-Tuning Image-Conditional Diffusion Models is Easier than You Think},
  author    = {Martin Garcia, Gonzalo and Abou Zeid, Karim and Schmidt, Christian and de Geus, Daan and Hermans, Alexander and Leibe, Bastian},
  booktitle = {Proceedings of the IEEE/CVF Winter Conference on Applications of Computer Vision (WACV)},
  year      = {2025}
}

@inproceedings{zhu2017unpaired,
  title={Unpaired image-to-image translation using cycle-consistent adversarial networks},
  author={Zhu, Jun-Yan and Park, Taesung and Isola, Phillip and Efros, Alexei A},
  booktitle={Proceedings of the IEEE international conference on computer vision},
  pages={2223--2232},
  year={2017}
}

@article{salimans2022progressive,
  title={Progressive distillation for fast sampling of diffusion models},
  author={Salimans, Tim and Ho, Jonathan},
  journal={arXiv preprint arXiv:2202.00512},
  year={2022}
}

@article{kingma2013auto,
  title={Auto-encoding variational bayes},
  author={Kingma, Diederik P},
  journal={arXiv preprint arXiv:1312.6114},
  year={2013}
}

@article{ho2020denoising,
  title={Denoising diffusion probabilistic models},
  author={Ho, Jonathan and Jain, Ajay and Abbeel, Pieter},
  journal={Advances in neural information processing systems},
  volume={33},
  pages={6840--6851},
  year={2020}
}

@article{ho2022classifier,
  title={Classifier-free diffusion guidance},
  author={Ho, Jonathan and Salimans, Tim},
  journal={arXiv preprint arXiv:2207.12598},
  year={2022}
}

@article{ho2022video,
  title={Video diffusion models},
  author={Ho, Jonathan and Salimans, Tim and Gritsenko, Alexey and Chan, William and Norouzi, Mohammad and Fleet, David J},
  journal={Advances in Neural Information Processing Systems},
  volume={35},
  pages={8633--8646},
  year={2022}
}

@inproceedings{
cong2024flatten,
title={{FLATTEN}: optical {FL}ow-guided {ATTEN}tion for consistent text-to-video editing},
author={Yuren Cong and Mengmeng Xu and christian simon and Shoufa Chen and Jiawei Ren and Yanping Xie and Juan-Manuel Perez-Rua and Bodo Rosenhahn and Tao Xiang and Sen He},
booktitle={The Twelfth International Conference on Learning Representations},
year={2024},
url={https://openreview.net/forum?id=JgqftqZQZ7}
}

@inproceedings{rombach2022high,
  title={High-resolution image synthesis with latent diffusion models},
  author={Rombach, Robin and Blattmann, Andreas and Lorenz, Dominik and Esser, Patrick and Ommer, Bj{\"o}rn},
  booktitle={Proceedings of the IEEE/CVF conference on computer vision and pattern recognition},
  pages={10684--10695},
  year={2022}
}

@article{saharia2022photorealistic,
  title={Photorealistic text-to-image diffusion models with deep language understanding},
  author={Saharia, Chitwan and Chan, William and Saxena, Saurabh and Li, Lala and Whang, Jay and Denton, Emily L and Ghasemipour, Kamyar and Gontijo Lopes, Raphael and Karagol Ayan, Burcu and Salimans, Tim and others},
  journal={Advances in neural information processing systems},
  volume={35},
  pages={36479--36494},
  year={2022}
}

@inproceedings{zhang2023adding,
  title={Adding conditional control to text-to-image diffusion models},
  author={Zhang, Lvmin and Rao, Anyi and Agrawala, Maneesh},
  booktitle={Proceedings of the IEEE/CVF international conference on computer vision},
  pages={3836--3847},
  year={2023}
}

@inproceedings{esser2024scaling,
  title={Scaling rectified flow transformers for high-resolution image synthesis},
  author={Esser, Patrick and Kulal, Sumith and Blattmann, Andreas and Entezari, Rahim and M{\"u}ller, Jonas and Saini, Harry and Levi, Yam and Lorenz, Dominik and Sauer, Axel and Boesel, Frederic and others},
  booktitle={Forty-first international conference on machine learning},
  year={2024}
}

@article{blattmann2023stable,
  title={Stable video diffusion: Scaling latent video diffusion models to large datasets},
  author={Blattmann, Andreas and Dockhorn, Tim and Kulal, Sumith and Mendelevitch, Daniel and Kilian, Maciej and Lorenz, Dominik and Levi, Yam and English, Zion and Voleti, Vikram and Letts, Adam and others},
  journal={arXiv preprint arXiv:2311.15127},
  year={2023}
}

@article{ye2023ip,
  title={Ip-adapter: Text compatible image prompt adapter for text-to-image diffusion models},
  author={Ye, Hu and Zhang, Jun and Liu, Sibo and Han, Xiao and Yang, Wei},
  journal={arXiv preprint arXiv:2308.06721},
  year={2023}
}

@inproceedings{mou2024t2i,
  title={T2i-adapter: Learning adapters to dig out more controllable ability for text-to-image diffusion models},
  author={Mou, Chong and Wang, Xintao and Xie, Liangbin and Wu, Yanze and Zhang, Jian and Qi, Zhongang and Shan, Ying},
  booktitle={Proceedings of the AAAI conference on artificial intelligence},
  volume={38},
  number={5},
  pages={4296--4304},
  year={2024}
}

@article{qin2023unicontrol,
  title={Unicontrol: A unified diffusion model for controllable visual generation in the wild},
  author={Qin, Can and Zhang, Shu and Yu, Ning and Feng, Yihao and Yang, Xinyi and Zhou, Yingbo and Wang, Huan and Niebles, Juan Carlos and Xiong, Caiming and Savarese, Silvio and others},
  journal={arXiv preprint arXiv:2305.11147},
  year={2023}
}

@inproceedings{brooks2023instructpix2pix,
  title={Instructpix2pix: Learning to follow image editing instructions},
  author={Brooks, Tim and Holynski, Aleksander and Efros, Alexei A},
  booktitle={Proceedings of the IEEE/CVF conference on computer vision and pattern recognition},
  pages={18392--18402},
  year={2023}
}

@inproceedings{kawar2023imagic,
  title={Imagic: Text-based real image editing with diffusion models},
  author={Kawar, Bahjat and Zada, Shiran and Lang, Oran and Tov, Omer and Chang, Huiwen and Dekel, Tali and Mosseri, Inbar and Irani, Michal},
  booktitle={Proceedings of the IEEE/CVF conference on computer vision and pattern recognition},
  pages={6007--6017},
  year={2023}
}

@inproceedings{yang2023paint,
  title={Paint by example: Exemplar-based image editing with diffusion models},
  author={Yang, Binxin and Gu, Shuyang and Zhang, Bo and Zhang, Ting and Chen, Xuejin and Sun, Xiaoyan and Chen, Dong and Wen, Fang},
  booktitle={Proceedings of the IEEE/CVF conference on computer vision and pattern recognition},
  pages={18381--18391},
  year={2023}
}

@inproceedings{chen2024anydoor,
  title={Anydoor: Zero-shot object-level image customization},
  author={Chen, Xi and Huang, Lianghua and Liu, Yu and Shen, Yujun and Zhao, Deli and Zhao, Hengshuang},
  booktitle={Proceedings of the IEEE/CVF conference on computer vision and pattern recognition},
  pages={6593--6602},
  year={2024}
}

@article{wang2024autostory,
  title={AutoStory: Generating Diverse Storytelling Images with Minimal Human Efforts},
  author={Wang, Wen and Zhao, Canyu and Chen, Hao and Chen, Zhekai and Zheng, Kecheng and Shen, Chunhua},
  journal={International Journal of Computer Vision},
  pages={1--22},
  year={2024},
  publisher={Springer}
}

@article{zhou2025storydiffusion,
  title={Storydiffusion: Consistent self-attention for long-range image and video generation},
  author={Zhou, Yupeng and Zhou, Daquan and Cheng, Ming-Ming and Feng, Jiashi and Hou, Qibin},
  journal={Advances in Neural Information Processing Systems},
  volume={37},
  pages={110315--110340},
  year={2025}
}

@inproceedings{ceylan2023pix2video,
  title={Pix2video: Video editing using image diffusion},
  author={Ceylan, Duygu and Huang, Chun-Hao P and Mitra, Niloy J},
  booktitle={Proceedings of the IEEE/CVF International Conference on Computer Vision},
  pages={23206--23217},
  year={2023}
}

@inproceedings{liu2024video,
  title={Video-p2p: Video editing with cross-attention control},
  author={Liu, Shaoteng and Zhang, Yuechen and Li, Wenbo and Lin, Zhe and Jia, Jiaya},
  booktitle={Proceedings of the IEEE/CVF Conference on Computer Vision and Pattern Recognition},
  pages={8599--8608},
  year={2024}
}

@inproceedings{chai2023stablevideo,
  title={Stablevideo: Text-driven consistency-aware diffusion video editing},
  author={Chai, Wenhao and Guo, Xun and Wang, Gaoang and Lu, Yan},
  booktitle={Proceedings of the IEEE/CVF International Conference on Computer Vision},
  pages={23040--23050},
  year={2023}
}

@inproceedings{ke2024repurposing,
  title={Repurposing diffusion-based image generators for monocular depth estimation},
  author={Ke, Bingxin and Obukhov, Anton and Huang, Shengyu and Metzger, Nando and Daudt, Rodrigo Caye and Schindler, Konrad},
  booktitle={Proceedings of the IEEE/CVF Conference on Computer Vision and Pattern Recognition},
  pages={9492--9502},
  year={2024}
}

@inproceedings{fu2024geowizard,
  title={Geowizard: Unleashing the diffusion priors for 3d geometry estimation from a single image},
  author={Fu, Xiao and Yin, Wei and Hu, Mu and Wang, Kaixuan and Ma, Yuexin and Tan, Ping and Shen, Shaojie and Lin, Dahua and Long, Xiaoxiao},
  booktitle={European Conference on Computer Vision},
  pages={241--258},
  year={2024},
  organization={Springer}
}

@article{gui2024depthfm,
  title={Depthfm: Fast monocular depth estimation with flow matching},
  author={Gui, Ming and Schusterbauer, Johannes and Prestel, Ulrich and Ma, Pingchuan and Kotovenko, Dmytro and Grebenkova, Olga and Baumann, Stefan Andreas and Hu, Vincent Tao and Ommer, Bj{\"o}rn},
  journal={arXiv preprint arXiv:2403.13788},
  year={2024}
}

@article{ye2024stablenormal,
  title={Stablenormal: Reducing diffusion variance for stable and sharp normal},
  author={Ye, Chongjie and Qiu, Lingteng and Gu, Xiaodong and Zuo, Qi and Wu, Yushuang and Dong, Zilong and Bo, Liefeng and Xiu, Yuliang and Han, Xiaoguang},
  journal={ACM Transactions on Graphics (TOG)},
  volume={43},
  number={6},
  pages={1--18},
  year={2024},
  publisher={ACM New York, NY, USA}
}

@article{xu2024matters,
  title={What Matters When Repurposing Diffusion Models for General Dense Perception Tasks?},
  author={Xu, Guangkai and Ge, Yongtao and Liu, Mingyu and Fan, Chengxiang and Xie, Kangyang and Zhao, Zhiyue and Chen, Hao and Shen, Chunhua},
  journal={arXiv preprint arXiv:2403.06090},
  year={2024}
}

@article{zhu2024unleashing,
  title={Unleashing the potential of the diffusion model in few-shot semantic segmentation},
  author={Zhu, Muzhi and Liu, Yang and Luo, Zekai and Jing, Chenchen and Chen, Hao and Xu, Guangkai and Wang, Xinlong and Shen, Chunhua},
  journal={arXiv preprint arXiv:2410.02369},
  year={2024}
}

@article{he2024lotus,
  title={Lotus: Diffusion-based visual foundation model for high-quality dense prediction},
  author={He, Jing and Li, Haodong and Yin, Wei and Liang, Yixun and Li, Leheng and Zhou, Kaiqiang and Zhang, Hongbo and Liu, Bingbing and Chen, Ying-Cong},
  journal={arXiv preprint arXiv:2409.18124},
  year={2024}
}

@inproceedings{gong2023diffpose,
  title={Diffpose: Toward more reliable 3d pose estimation},
  author={Gong, Jia and Foo, Lin Geng and Fan, Zhipeng and Ke, Qiuhong and Rahmani, Hossein and Liu, Jun},
  booktitle={Proceedings of the IEEE/CVF Conference on Computer Vision and Pattern Recognition},
  pages={13041--13051},
  year={2023}
}

@article{wang2024lavin,
  title={Lavin-dit: Large vision diffusion transformer},
  author={Wang, Zhaoqing and Xia, Xiaobo and Chen, Runnan and Yu, Dongdong and Wang, Changhu and Gong, Mingming and Liu, Tongliang},
  journal={arXiv preprint arXiv:2411.11505},
  year={2024}
}

@article{le2024one,
  title={One Diffusion to Generate Them All},
  author={Le, Duong H and Pham, Tuan and Lee, Sangho and Clark, Christopher and Kembhavi, Aniruddha and Mandt, Stephan and Krishna, Ranjay and Lu, Jiasen},
  journal={arXiv preprint arXiv:2411.16318},
  year={2024}
}

@inproceedings{zeng2024dilightnet,
  title={Dilightnet: Fine-grained lighting control for diffusion-based image generation},
  author={Zeng, Chong and Dong, Yue and Peers, Pieter and Kong, Youkang and Wu, Hongzhi and Tong, Xin},
  booktitle={ACM SIGGRAPH 2024 Conference Papers},
  pages={1--12},
  year={2024}
}

@inproceedings{kocsis2024lightit,
  title={Lightit: Illumination modeling and control for diffusion models},
  author={Kocsis, Peter and Philip, Julien and Sunkavalli, Kalyan and Nie{\ss}ner, Matthias and Hold-Geoffroy, Yannick},
  booktitle={Proceedings of the IEEE/CVF Conference on Computer Vision and Pattern Recognition},
  pages={9359--9369},
  year={2024}
}

@inproceedings{ren2024relightful,
  title={Relightful harmonization: Lighting-aware portrait background replacement},
  author={Ren, Mengwei and Xiong, Wei and Yoon, Jae Shin and Shu, Zhixin and Zhang, Jianming and Jung, HyunJoon and Gerig, Guido and Zhang, He},
  booktitle={Proceedings of the IEEE/CVF Conference on Computer Vision and Pattern Recognition},
  pages={6452--6462},
  year={2024}
}

@article{jin2025neural,
  title={Neural gaffer: Relighting any object via diffusion},
  author={Jin, Haian and Li, Yuan and Luan, Fujun and Xiangli, Yuanbo and Bi, Sai and Zhang, Kai and Xu, Zexiang and Sun, Jin and Snavely, Noah},
  journal={Advances in Neural Information Processing Systems},
  volume={37},
  pages={141129--141152},
  year={2025}
}

@inproceedings{zhangscaling,
  title={Scaling In-the-Wild Training for Diffusion-based Illumination Harmonization and Editing by Imposing Consistent Light Transport},
  author={Zhang, Lvmin and Rao, Anyi and Agrawala, Maneesh},
  booktitle={The Thirteenth International Conference on Learning Representations},
  year={2025}
}

@inproceedings{kovacs2017shading,
  title={Shading annotations in the wild},
  author={Kovacs, Balazs and Bell, Sean and Snavely, Noah and Bala, Kavita},
  booktitle={Proceedings of the IEEE conference on computer vision and pattern recognition},
  pages={6998--7007},
  year={2017}
}

@inproceedings{wu2023measured,
  title={Measured albedo in the wild: Filling the gap in intrinsics evaluation},
  author={Wu, Jiaye and Chowdhury, Sanjoy and Shanmugaraja, Hariharmano and Jacobs, David and Sengupta, Soumyadip},
  booktitle={2023 IEEE International Conference on Computational Photography (ICCP)},
  pages={1--12},
  year={2023},
  organization={IEEE}
}

@inproceedings{zhou2015learning,
  title={Learning data-driven reflectance priors for intrinsic image decomposition},
  author={Zhou, Tinghui and Krahenbuhl, Philipp and Efros, Alexei A},
  booktitle={Proceedings of the IEEE international conference on computer vision},
  pages={3469--3477},
  year={2015}
}

@article{careaga2023intrinsic,
  title={Intrinsic image decomposition via ordinal shading},
  author={Careaga, Chris and Aksoy, Ya{\u{g}}{\i}z},
  journal={ACM Transactions on Graphics},
  volume={43},
  number={1},
  pages={1--24},
  year={2023},
  publisher={ACM New York, NY, USA}
}

@inproceedings{zoran2015learning,
  title={Learning ordinal relationships for mid-level vision},
  author={Zoran, Daniel and Isola, Phillip and Krishnan, Dilip and Freeman, William T},
  booktitle={Proceedings of the IEEE international conference on computer vision},
  pages={388--396},
  year={2015}
}

@inproceedings{ramanagopal2024theory,
  title={A theory of joint light and heat transport for lambertian scenes},
  author={Ramanagopal, Mani and Narayanan, Sriram and Sankaranarayanan, Aswin C and Narasimhan, Srinivasa G},
  booktitle={Proceedings of the IEEE/CVF Conference on Computer Vision and Pattern Recognition},
  pages={11924--11933},
  year={2024}
}

@inproceedings{yoshida2023light,
  title={Light source separation and intrinsic image decomposition under ac illumination},
  author={Yoshida, Yusaku and Kawahara, Ryo and Okabe, Takahiro},
  booktitle={Proceedings of the IEEE/CVF Conference on Computer Vision and Pattern Recognition},
  pages={5735--5743},
  year={2023}
}

@inproceedings{geng2023tree,
  title={Tree-structured shading decomposition},
  author={Geng, Chen and Yu, Hong-Xing and Zhang, Sharon and Agrawala, Maneesh and Wu, Jiajun},
  booktitle={Proceedings of the IEEE/CVF International Conference on Computer Vision},
  pages={488--498},
  year={2023}
}

@article{philip2019multi,
  title={Multi-view relighting using a geometry-aware network.},
  author={Philip, Julien and Gharbi, Micha{\"e}l and Zhou, Tinghui and Efros, Alexei A and Drettakis, George},
  journal={ACM Trans. Graph.},
  volume={38},
  number={4},
  pages={78--1},
  year={2019}
}

@inproceedings{ye2023intrinsicnerf,
  title={Intrinsicnerf: Learning intrinsic neural radiance fields for editable novel view synthesis},
  author={Ye, Weicai and Chen, Shuo and Bao, Chong and Bao, Hujun and Pollefeys, Marc and Cui, Zhaopeng and Zhang, Guofeng},
  booktitle={Proceedings of the IEEE/CVF International Conference on Computer Vision},
  pages={339--351},
  year={2023}
}

@article{wu2024deferredgs,
  title={Deferredgs: Decoupled and editable gaussian splatting with deferred shading},
  author={Wu, Tong and Sun, Jia-Mu and Lai, Yu-Kun and Ma, Yuewen and Kobbelt, Leif and Gao, Lin},
  journal={arXiv preprint arXiv:2404.09412},
  year={2024}
}

@inproceedings{karras2019style,
  title={A style-based generator architecture for generative adversarial networks},
  author={Karras, Tero and Laine, Samuli and Aila, Timo},
  booktitle={Proceedings of the IEEE/CVF conference on computer vision and pattern recognition},
  pages={4401--4410},
  year={2019}
}

@article{bhattad2023stylegan,
  title={Stylegan knows normal, depth, albedo, and more},
  author={Bhattad, Anand and McKee, Daniel and Hoiem, Derek and Forsyth, David},
  journal={Advances in Neural Information Processing Systems},
  volume={36},
  pages={73082--73103},
  year={2023}
}

@article{du2023generative,
  title={Generative models: What do they know? do they know things? let's find out!},
  author={Du, Xiaodan and Kolkin, Nicholas and Shakhnarovich, Greg and Bhattad, Anand},
  journal={arXiv preprint arXiv:2311.17137},
  year={2023}
}

@inproceedings{luo2024intrinsicdiffusion,
  title={Intrinsicdiffusion: Joint intrinsic layers from latent diffusion models},
  author={Luo, Jundan and Ceylan, Duygu and Yoon, Jae Shin and Zhao, Nanxuan and Philip, Julien and Fr{\"u}hst{\"u}ck, Anna and Li, Wenbin and Richardt, Christian and Wang, Tuanfeng},
  booktitle={ACM SIGGRAPH 2024 Conference Papers},
  pages={1--11},
  year={2024}
}

@inproceedings{li2021openrooms,
  title={Openrooms: An open framework for photorealistic indoor scene datasets},
  author={Li, Zhengqin and Yu, Ting-Wei and Sang, Shen and Wang, Sarah and Song, Meng and Liu, Yuhan and Yeh, Yu-Ying and Zhu, Rui and Gundavarapu, Nitesh and Shi, Jia and others},
  booktitle={Proceedings of the IEEE/CVF conference on computer vision and pattern recognition},
  pages={7190--7199},
  year={2021}
}

@inproceedings{zhu2022learning,
  title={Learning-based inverse rendering of complex indoor scenes with differentiable Monte Carlo raytracing},
  author={Zhu, Jingsen and Luan, Fujun and Huo, Yuchi and Lin, Zihao and Zhong, Zhihua and Xi, Dianbing and Wang, Rui and Bao, Hujun and Zheng, Jiaxiang and Tang, Rui},
  booktitle={SIGGRAPH Asia 2022 Conference Papers},
  pages={1--8},
  year={2022}
}

@inproceedings{li2023matrixcity,
  title={Matrixcity: A large-scale city dataset for city-scale neural rendering and beyond},
  author={Li, Yixuan and Jiang, Lihan and Xu, Linning and Xiangli, Yuanbo and Wang, Zhenzhi and Lin, Dahua and Dai, Bo},
  booktitle={Proceedings of the IEEE/CVF International Conference on Computer Vision},
  pages={3205--3215},
  year={2023}
}

@inproceedings{roberts2021hypersim,
  title={Hypersim: A photorealistic synthetic dataset for holistic indoor scene understanding},
  author={Roberts, Mike and Ramapuram, Jason and Ranjan, Anurag and Kumar, Atulit and Bautista, Miguel Angel and Paczan, Nathan and Webb, Russ and Susskind, Joshua M},
  booktitle={Proceedings of the IEEE/CVF international conference on computer vision},
  pages={10912--10922},
  year={2021}
}

@inproceedings {InteriorNet18,
      author = { Wenbin Li and Sajad Saeedi and John McCormac and Ronald Clark and 
                 Dimos Tzoumanikas and Qing Ye and Yuzhong Huang and Rui Tang and 
                 Stefan Leutenegger },
   booktitle = { British Machine Vision Conference (BMVC) },
       title = { InteriorNet: Mega-scale Multi-sensor Photo-realistic Indoor Scenes Dataset },
        year = { 2018 }
}

@book{pharr2023physically,
  title={Physically based rendering: From theory to implementation},
  author={Pharr, Matt and Jakob, Wenzel and Humphreys, Greg},
  year={2023},
  publisher={MIT Press}
}

@inproceedings{nalbach2017deep,
  title={Deep shading: convolutional neural networks for screen space shading},
  author={Nalbach, Oliver and Arabadzhiyska, Elena and Mehta, Dushyant and Seidel, H-P and Ritschel, Tobias},
  booktitle={Computer graphics forum},
  volume={36},
  number={4},
  pages={65--78},
  year={2017},
  organization={Wiley Online Library}
}

@inproceedings{griffiths2022outcast,
  title={OutCast: Outdoor Single-image Relighting with Cast Shadows},
  author={Griffiths, David and Ritschel, Tobias and Philip, Julien},
  booktitle={Computer Graphics Forum},
  volume={41},
  number={2},
  pages={179--193},
  year={2022},
  organization={Wiley Online Library}
}

@article{pandey2021total,
  title={Total relighting: learning to relight portraits for background replacement.},
  author={Pandey, Rohit and Orts-Escolano, Sergio and Legendre, Chloe and Haene, Christian and Bouaziz, Sofien and Rhemann, Christoph and Debevec, Paul E and Fanello, Sean Ryan},
  journal={ACM Trans. Graph.},
  volume={40},
  number={4},
  pages={43--1},
  year={2021}
}

@inproceedings{yu2020self,
  title={Self-supervised outdoor scene relighting},
  author={Yu, Ye and Meka, Abhimitra and Elgharib, Mohamed and Seidel, Hans-Peter and Theobalt, Christian and Smith, William AP},
  booktitle={Computer Vision--ECCV 2020: 16th European Conference, Glasgow, UK, August 23--28, 2020, Proceedings, Part XXII 16},
  pages={84--101},
  year={2020},
  organization={Springer}
}

@inproceedings{kim2024switchlight,
  title={SwitchLight: Co-design of Physics-driven Architecture and Pre-training Framework for Human Portrait Relighting},
  author={Kim, Hoon and Jang, Minje and Yoon, Wonjun and Lee, Jisoo and Na, Donghyun and Woo, Sanghyun},
  booktitle={Proceedings of the IEEE/CVF Conference on Computer Vision and Pattern Recognition},
  pages={25096--25106},
  year={2024}
}

@inproceedings{rudnev2022nerf,
  title={Nerf for outdoor scene relighting},
  author={Rudnev, Viktor and Elgharib, Mohamed and Smith, William and Liu, Lingjie and Golyanik, Vladislav and Theobalt, Christian},
  booktitle={European Conference on Computer Vision},
  pages={615--631},
  year={2022},
  organization={Springer}
}

@inproceedings{wang2023neural,
  title={Neural fields meet explicit geometric representations for inverse rendering of urban scenes},
  author={Wang, Zian and Shen, Tianchang and Gao, Jun and Huang, Shengyu and Munkberg, Jacob and Hasselgren, Jon and Gojcic, Zan and Chen, Wenzheng and Fidler, Sanja},
  booktitle={Proceedings of the IEEE/CVF Conference on Computer Vision and Pattern Recognition},
  pages={8370--8380},
  year={2023}
}

@article{lin2023urbanir,
  title={Urbanir: Large-scale urban scene inverse rendering from a single video},
  author={Lin, Zhi-Hao and Liu, Bohan and Chen, Yi-Ting and Forsyth, David and Huang, Jia-Bin and Bhattad, Anand and Wang, Shenlong},
  journal={arXiv preprint arXiv:2306.09349},
  year={2023}
}

@article{zhang2024zerocomp,
  title={Zerocomp: Zero-shot object compositing from image intrinsics via diffusion},
  author={Zhang, Zitian and Fortier-Chouinard, Fr{\'e}d{\'e}ric and Garon, Mathieu and Bhattad, Anand and Lalonde, Jean-Fran{\c{c}}ois},
  journal={arXiv preprint arXiv:2410.08168},
  year={2024}
}

@inproceedings{ramamoorthi2001signal,
  title={A signal-processing framework for inverse rendering},
  author={Ramamoorthi, Ravi and Hanrahan, Pat},
  booktitle={Proceedings of the 28th annual conference on Computer graphics and interactive techniques},
  pages={117--128},
  year={2001}
}

@article{barron2014shape,
  title={Shape, illumination, and reflectance from shading},
  author={Barron, Jonathan T and Malik, Jitendra},
  journal={IEEE transactions on pattern analysis and machine intelligence},
  volume={37},
  number={8},
  pages={1670--1687},
  year={2014},
  publisher={IEEE}
}

@article{liang2025diffusionrenderer,
  title={DiffusionRenderer: Neural Inverse and Forward Rendering with Video Diffusion Models},
  author={Liang, Ruofan and Gojcic, Zan and Ling, Huan and Munkberg, Jacob and Hasselgren, Jon and Lin, Zhi-Hao and Gao, Jun and Keller, Alexander and Vijaykumar, Nandita and Fidler, Sanja and others},
  journal={arXiv preprint arXiv:2501.18590},
  year={2025}
}

@article{careaga2024colorful,
  title={Colorful diffuse intrinsic image decomposition in the wild},
  author={Careaga, Chris and Aksoy, Ya{\u{g}}{\i}z},
  journal={ACM Transactions on Graphics (TOG)},
  volume={43},
  number={6},
  pages={1--12},
  year={2024},
  publisher={ACM New York, NY, USA}
}

@inproceedings{controlnet_plus_plus,
    author    = {Ming Li and Taojiannan Yang and Huafeng Kuang and Jie Wu and Zhaoning Wang and Xuefeng Xiao and Chen Chen},
    title     = {ControlNet{++}: Improving Conditional Controls with Efficient Consistency Feedback},
    booktitle = {European Conference on Computer Vision},
    year      = {2024},
}

@inproceedings{chen2024intrinsicanything,
  title={Intrinsicanything: Learning diffusion priors for inverse rendering under unknown illumination},
  author={Chen, Xi and Peng, Sida and Yang, Dongchen and Liu, Yuan and Pan, Bowen and Lv, Chengfei and Zhou, Xiaowei},
  booktitle={European Conference on Computer Vision},
  pages={450--467},
  year={2024},
  organization={Springer}
}

@inproceedings{plummer2015flickr30k,
  title={Flickr30k entities: Collecting region-to-phrase correspondences for richer image-to-sentence models},
  author={Plummer, Bryan A and Wang, Liwei and Cervantes, Chris M and Caicedo, Juan C and Hockenmaier, Julia and Lazebnik, Svetlana},
  booktitle={Proceedings of the IEEE international conference on computer vision},
  pages={2641--2649},
  year={2015}
}

@inproceedings{lin2014microsoft,
  title={Microsoft coco: Common objects in context},
  author={Lin, Tsung-Yi and Maire, Michael and Belongie, Serge and Hays, James and Perona, Pietro and Ramanan, Deva and Doll{\'a}r, Piotr and Zitnick, C Lawrence},
  booktitle={Computer vision--ECCV 2014: 13th European conference, zurich, Switzerland, September 6-12, 2014, proceedings, part v 13},
  pages={740--755},
  year={2014},
  organization={Springer}
}

@inproceedings{li2022blip,
  title={Blip: Bootstrapping language-image pre-training for unified vision-language understanding and generation},
  author={Li, Junnan and Li, Dongxu and Xiong, Caiming and Hoi, Steven},
  booktitle={International conference on machine learning},
  pages={12888--12900},
  year={2022},
  organization={PMLR}
}

@inproceedings{li2023blip,
  title={Blip-2: Bootstrapping language-image pre-training with frozen image encoders and large language models},
  author={Li, Junnan and Li, Dongxu and Savarese, Silvio and Hoi, Steven},
  booktitle={International conference on machine learning},
  pages={19730--19742},
  year={2023},
  organization={PMLR}
}

@inproceedings{grosse2009ground,
  title={Ground truth dataset and baseline evaluations for intrinsic image algorithms},
  author={Grosse, Roger and Johnson, Micah K and Adelson, Edward H and Freeman, William T},
  booktitle={2009 IEEE 12th International Conference on Computer Vision},
  pages={2335--2342},
  year={2009},
  organization={IEEE}
}

@inproceedings{infinigen2023infinite,
  title={Infinite Photorealistic Worlds Using Procedural Generation},
  author={Raistrick, Alexander and Lipson, Lahav and Ma, Zeyu and Mei, Lingjie and Wang, Mingzhe and Zuo, Yiming and Kayan, Karhan and Wen, Hongyu and Han, Beining and Wang, Yihan and Newell, Alejandro and Law, Hei and Goyal, Ankit and Yang, Kaiyu and Deng, Jia},
  booktitle={Proceedings of the IEEE/CVF Conference on Computer Vision and Pattern Recognition},
  pages={12630--12641},
  year={2023}
}

@inproceedings{infinigen2024indoors,
    author    = {Raistrick, Alexander and Mei, Lingjie and Kayan, Karhan and Yan, David and Zuo, Yiming and Han, Beining and Wen, Hongyu and Parakh, Meenal and Alexandropoulos, Stamatis and Lipson, Lahav and Ma, Zeyu and Deng, Jia},
    title     = {Infinigen Indoors: Photorealistic Indoor Scenes using Procedural Generation},
    booktitle = {Proceedings of the IEEE/CVF Conference on Computer Vision and Pattern Recognition (CVPR)},
    month     = {June},
    year      = {2024},
    pages     = {21783-21794}
}

@inproceedings{li2020inverse,
  title={Inverse rendering for complex indoor scenes: Shape, spatially-varying lighting and svbrdf from a single image},
  author={Li, Zhengqin and Shafiei, Mohammad and Ramamoorthi, Ravi and Sunkavalli, Kalyan and Chandraker, Manmohan},
  booktitle={Proceedings of the IEEE/CVF conference on computer vision and pattern recognition},
  pages={2475--2484},
  year={2020}
}

@inproceedings{zhu2022irisformer,
  title={Irisformer: Dense vision transformers for single-image inverse rendering in indoor scenes},
  author={Zhu, Rui and Li, Zhengqin and Matai, Janarbek and Porikli, Fatih and Chandraker, Manmohan},
  booktitle={Proceedings of the IEEE/CVF Conference on Computer Vision and Pattern Recognition},
  pages={2822--2831},
  year={2022}
}

@article{ranftl2020towards,
  title={Towards robust monocular depth estimation: Mixing datasets for zero-shot cross-dataset transfer},
  author={Ranftl, Ren{\'e} and Lasinger, Katrin and Hafner, David and Schindler, Konrad and Koltun, Vladlen},
  journal={IEEE transactions on pattern analysis and machine intelligence},
  volume={44},
  number={3},
  pages={1623--1637},
  year={2020},
  publisher={IEEE}
}
}


\end{document}